\documentclass[runningheads]{llncs}
\usepackage{graphicx}

\usepackage{tikz}
\usepackage{comment}
\usepackage{amssymb}
\usepackage[fleqn]{amsmath}
\usepackage{color}
\usepackage{multirow}
\usepackage{xspace}
\usepackage{booktabs}
\usepackage[accsupp]{axessibility} 
\usepackage{caption} 
\usepackage{orcidlink}
\usepackage{adjustbox}
\usepackage[figuresright]{rotating}

\newcommand{\textBF}[1]{%
    \pdfliteral direct {2 Tr 0.6 w} %
     #1%
    \pdfliteral direct {0 Tr 0 w}%
}

\usepackage{hyperref}
\hypersetup{
  colorlinks   = true,    %
  urlcolor     = blue,    %
  linkcolor    = blue,    %
  citecolor    = green    %
}

\begin{document}
\pagestyle{headings}
\mainmatter
\def\ECCVSubNumber{1342}

\title{StretchBEV: Stretching Future Instance Prediction Spatially and Temporally}

\author{Adil Kaan Akan\orcidlink{0000-0003-0022-7224}\index{Akan, Adil Kaan} \and
Fatma G\"{u}ney\orcidlink{0000-0002-0358-983X}}
\authorrunning{Akan and G\"{u}ney}
\titlerunning{StretchBEV}
\institute{KUIS AI Center, Koc University,
Istanbul, Turkey\\
\email{\{kakan20, fguney\}@ku.edu.tr}\\
\url{https://kuis-ai.github.io/stretchbev}}

\maketitle

\newcommand{\Perp}{\perp\!\!\! \perp}
\newcommand{\bK}{\mathbf{K}}
\newcommand{\bX}{\mathbf{X}}
\newcommand{\bY}{\mathbf{Y}}
\newcommand{\bk}{\mathbf{k}}
\newcommand{\bx}{\mathbf{x}}
\newcommand{\by}{\mathbf{y}}
\newcommand{\bhy}{\hat{\mathbf{y}}}
\newcommand{\bty}{\tilde{\mathbf{y}}}
\newcommand{\bG}{\mathbf{G}}
\newcommand{\bI}{\mathbf{I}}
\newcommand{\bg}{\mathbf{g}}
\newcommand{\bS}{\mathbf{S}}
\newcommand{\bs}{\mathbf{s}}
\newcommand{\bM}{\mathbf{M}}
\newcommand{\bw}{\mathbf{w}}
\newcommand{\eye}{\mathbf{I}}
\newcommand{\bU}{\mathbf{U}}
\newcommand{\bV}{\mathbf{V}}
\newcommand{\bW}{\mathbf{W}}
\newcommand{\bn}{\mathbf{n}}
\newcommand{\bv}{\mathbf{v}}
\newcommand{\bwv}{\mathbf{wv}}
\newcommand{\bq}{\mathbf{q}}
\newcommand{\bR}{\mathbf{R}}
\newcommand{\bi}{\mathbf{i}}
\newcommand{\bj}{\mathbf{j}}
\newcommand{\bp}{\mathbf{p}}
\newcommand{\bt}{\mathbf{t}}
\newcommand{\bJ}{\mathbf{J}}
\newcommand{\bu}{\mathbf{u}}
\newcommand{\bB}{\mathbf{B}}
\newcommand{\bD}{\mathbf{D}}
\newcommand{\bz}{\mathbf{z}}
\newcommand{\bP}{\mathbf{P}}
\newcommand{\bC}{\mathbf{C}}
\newcommand{\bA}{\mathbf{A}}
\newcommand{\bZ}{\mathbf{Z}}
\newcommand{\bff}{\mathbf{f}}
\newcommand{\bF}{\mathbf{F}}
\newcommand{\bo}{\mathbf{o}}
\newcommand{\bO}{\mathbf{O}}
\newcommand{\bc}{\mathbf{c}}
\newcommand{\bm}{\mathbf{m}}
\newcommand{\bT}{\mathbf{T}}
\newcommand{\bQ}{\mathbf{Q}}
\newcommand{\bL}{\mathbf{L}}
\newcommand{\bl}{\mathbf{l}}
\newcommand{\ba}{\mathbf{a}}
\newcommand{\bE}{\mathbf{E}}
\newcommand{\bH}{\mathbf{H}}
\newcommand{\bd}{\mathbf{d}}
\newcommand{\br}{\mathbf{r}}
\newcommand{\be}{\mathbf{e}}
\newcommand{\bb}{\mathbf{b}}
\newcommand{\bh}{\mathbf{h}}
\newcommand{\bhh}{\hat{\mathbf{h}}}
\newcommand{\btheta}{\boldsymbol{\theta}}
\newcommand{\bTheta}{\boldsymbol{\Theta}}
\newcommand{\bpi}{\boldsymbol{\pi}}
\newcommand{\bphi}{\boldsymbol{\phi}}
\newcommand{\bpsi}{\boldsymbol{\psi}}
\newcommand{\bPhi}{\boldsymbol{\Phi}}
\newcommand{\bmu}{\boldsymbol{\mu}}
\newcommand{\bsigma}{\boldsymbol{\sigma}}
\newcommand{\bSigma}{\boldsymbol{\Sigma}}
\newcommand{\bGamma}{\boldsymbol{\Gamma}}
\newcommand{\bbeta}{\boldsymbol{\beta}}
\newcommand{\bomega}{\boldsymbol{\omega}}
\newcommand{\blambda}{\boldsymbol{\lambda}}
\newcommand{\bLambda}{\boldsymbol{\Lambda}}
\newcommand{\bkappa}{\boldsymbol{\kappa}}
\newcommand{\btau}{\boldsymbol{\tau}}
\newcommand{\balpha}{\boldsymbol{\alpha}}
\newcommand{\nR}{\mathbb{R}}
\newcommand{\nN}{\mathbb{N}}
\newcommand{\nL}{\mathbb{L}}
\newcommand{\nE}{\mathbb{E}}
\newcommand{\cN}{\mathcal{N}}
\newcommand{\cM}{\mathcal{M}}
\newcommand{\cR}{\mathcal{R}}
\newcommand{\cB}{\mathcal{B}}
\newcommand{\cL}{\mathcal{L}}
\newcommand{\cH}{\mathcal{H}}
\newcommand{\cS}{\mathcal{S}}
\newcommand{\cT}{\mathcal{T}}
\newcommand{\cO}{\mathcal{O}}
\newcommand{\cC}{\mathcal{C}}
\newcommand{\cP}{\mathcal{P}}
\newcommand{\cE}{\mathcal{E}}
\newcommand{\cI}{\mathcal{I}}
\newcommand{\cF}{\mathcal{F}}
\newcommand{\cK}{\mathcal{K}}
\newcommand{\cY}{\mathcal{Y}}
\newcommand{\cX}{\mathcal{X}}
\def\bgamma{\boldsymbol\gamma}

\newcommand{\specialcell}[2][c]{%
  \begin{tabular}[#1]{@{}c@{}}#2\end{tabular}}

\newcommand{\figref}[1]{\Fig~\ref{#1}}
\newcommand{\secref}[1]{Section~\ref{#1}}
\newcommand{\algref}[1]{Algorithm~\ref{#1}}
\newcommand{\eqnref}[1]{Eq.~\eqref{#1}}
\newcommand{\tabref}[1]{Table~\ref{#1}}

\newcommand{\rulesep}{\unskip\ \vrule\ }

\newcommand{\KLD}[2]{D_{\mathrm{KL}} \Big(#1 \mid\mid #2 \Big)}

\renewcommand{\b}{\ensuremath{\mathbf}}

\def\mc{\mathcal}
\def\mb{\mathbf}

\newcommand{\T}{^{\raisemath{-1pt}{\mathsf{T}}}}

\makeatletter
\DeclareRobustCommand\onedot{\futurelet\@let@token\@onedot}
\def\@onedot{\ifx\@let@token.\else.\null\fi\xspace}
\def\eg{e.g\onedot} \def\Eg{E.g\onedot}
\def\ie{i.e\onedot} \def\Ie{I.e\onedot}
\def\cf{cf\onedot} \def\Cf{Cf\onedot}
\def\etc{etc\onedot} \def\vs{vs\onedot}
\def\wrt{wrt\onedot}
\def\dof{d.o.f\onedot}
\def\etal{et~al\onedot} \def\iid{i.i.d\onedot}
\def\Fig{Fig\onedot} \def\Eqn{Eqn\onedot} \def\Sec{Sec\onedot} \def\Alg{Alg\onedot}
\makeatother

\newcommand{\xdownarrow}[1]{%
  {\left\downarrow\vbox to #1{}\right.\kern-\nulldelimiterspace}
}

\newcommand{\xuparrow}[1]{%
  {\left\uparrow\vbox to #1{}\right.\kern-\nulldelimiterspace}
}

\newcommand*\rot{\rotatebox{90}}
\newcommand{\boldparagraph}[1]{\vspace{0.15cm}\noindent{\bf #1:} }
\newcommand{\boldquestion}[1]{\vspace{0.2cm}\noindent{\bf #1} }

\newcommand{\ka}[1]{ \noindent {\color{blue} {\bf Kaan:} {#1}} } 
\newcommand{\ftm}[1]{ \noindent {\color{magenta} {\bf Fatma:} {#1}} }
\begin{abstract}
In self-driving, predicting future in terms of location and motion of all the agents around the vehicle is a crucial requirement for planning. Recently, a new joint formulation of perception and prediction has emerged by fusing rich sensory information perceived from multiple cameras into a compact bird's-eye view representation to perform prediction. However, the quality of future predictions degrades over time while extending to longer time horizons due to multiple plausible predictions. In this work, we address this inherent uncertainty in future predictions with a stochastic temporal model. Our model learns temporal dynamics in a latent space through stochastic residual updates at each time step. By sampling from a learned distribution at each time step, we obtain more diverse future predictions that are also more accurate compared to previous work, especially stretching both spatially further regions in the scene and temporally over longer time horizons. Despite separate processing of each time step, our model is still efficient through decoupling of the learning of dynamics and the generation of future predictions.
\end{abstract}
\section{Introduction}
Future prediction with sequential visual data has been studied from different perspectives. Stochastic video prediction methods operate in the pixel space and learn to predict future frames conditioned on the previous frames. These methods achieve impressive results by modelling the uncertainty of the future with stochasticity, however, long-term predictions in real-world sequences tend to be quite blurry due to the complexity of directly predicting pixels. A more practical approach is to consider a compact representation that is tightly connected to the modalities required for the downstream application. In self-driving, understanding the 3D properties of the scene and the motion of other agents plays a key role in future predictions. The bird's-eye view~(BEV) representation meets these requirements by first fusing information from multiple cameras into a 3D point cloud and then projecting the points to the ground plane~\cite{Philion2020ECCV}. This leads to a compact representation where the future location and motion of multiple agents in the scene can be reliably predicted. In this paper, we explore the potential of stochastic future prediction for self-driving to produce admissible and diverse results in long sequences with an efficient and compact BEV representation. 

Future prediction from the BEV representation has been recently proposed in FIERY~\cite{Hu2021ICCV}. The BEV representations of past frames are first related in a temporal model and then used to learn two distributions representing the present and the future. Based on sampling from one of these distributions depending on train or test time, various future modalities are predicted with a recurrent model. For planning, long term multiple future predictions are crucial, however, the predictions of FIERY degrade over longer time spans due to the limited representation capability of a single distribution for increasing diversity in longer predictions. Following the official implementation, the predictions do not differ significantly based on random samples but converge to the mean, therefore lack diversity. We start from the same BEV representation and predict the same output modalities to be comparable. Differently, instead of two distributions for the present and the future, we propose to learn time-dependent distributions by predicting a residual change at each time step to better capture long-term dynamics. Furthermore, we show that by sampling a random variable at each time step, we can increase the diversity of future predictions while still being accurate and efficient. For efficiency, we use a state-space model \cite{Murphy2023Prob} instead of costly auto-regressive models.

The main idea behind the efficiency of the state-space model is to decouple the learning of dynamics and the generation of predictions~\cite{Franceschi2020ICML}. We first learn a low dimensional latent space from the BEV representation to capture the dynamics and then learn to decode some output modalities from the predictions in that latent space. These output modalities show the location and the motion of the agents in the scene.
We can train our dynamics model independently by learning to match the BEV representations of future frames that are predicted by our model to the ones that are extracted from a pre-trained BEV segmentation model~\cite{Philion2020ECCV}. Through residual updates to the latent space, our model can capture the changes to the BEV representation over time. This way, the only information we use from the future is the BEV representation of future frames.
Another option is to encode the future modalities to predict and provide the model with this encoded representation to learn a future distribution \cite{Hu2021ICCV}. We experiment with both options in this paper. While providing labels in the future distribution leads to better performance, learning the dynamics becomes dependent on the labels. From the BEV predictions, we train a decoder to predict the location and the motion of future instances in a supervised manner. These output modalities increase the interpretability of the predicted BEV representations that can be used for learning a driving policy in future work.

We present a stochastic future instance prediction method in BEV from multiple cameras. We formulate the prediction task as learning temporal dynamics through stochastic residual updates to a latent representation of the state at each time step. Our model can generate diverse predictions which are interpretable through supervised decoding of the predictions.
Our proposed approach clearly outperforms the state of the art in various metrics evaluating the decoded predictions, especially by large margins in challenging cases of spatially far regions and temporally long spans. We also show increased  diversity in the predictions. %

\section{Related Work}
\subsection{Stochastic Future Prediction}
Stochastic future prediction has been mostly explored in the context of next frame prediction in videos. In stochastic video generation, the goal is to predict future frames conditioned on a few initial frames. Typically, the main focus is the diversity of future predictions with a large number of samples for future and the number of frames to be predicted is at least twice as many as the initial conditioning frames. Our work fits into the stochastic future prediction framework, producing long term, diverse predictions, however, we predict future in the BEV space instead of the noisy pixel space.

Most of the stochastic video prediction methods~\cite{Denton2018ICML,Babaeizadeh2018ICLR} use a recurrent neural network in an auto-regressive manner by feeding the generated predictions back to the model to predict future. The performance of auto-regressive methods can be improved by increasing the network capacity~\cite{Villegas2019NeurIPS} or introducing a hierarchy into the latent variables~\cite{Castrejon2019ICCV}, which also increase the complexity of these methods. Due to complexity of predicting pixels, another group of work moves away from the pixel space to the keypoints~\cite{Minderer2019NeurIPS} or to the motion space by incorporating motion history~\cite{Akan2021ICCV}. Our proposed approach follows a similar strategy by performing future prediction in the BEV representation, but more efficiently by avoiding auto-regressive predictions.

Auto-regressive strategy requires encoding the predictions, leading to high computational cost and creates a tight coupling between the temporal dynamics and the generation process~\cite{Gregor2019ICLR,Rubanova2019NeurIPS}. The state-space models~(SSM) break this coupling by separating the learning of dynamics from the generation process, resulting in a computationally more efficient approach. Low-dimensional states still depend on previous states but not on predictions.
Furthermore, learned states can be directly used in reinforcement learning~\cite{Gregor2019ICLR} and can be interpreted~\cite{Rubanova2019NeurIPS}, making SSMs particularly appealing for self-driving. Earlier SSMs with variational deep inference suffer from complicated inference schemes and typically target low-dimensional data~\cite{Krishnan2017AAAI,Karl2017ICLR}. An efficient training strategy with a temporal model based on residual updates is proposed for high-dimensional video prediction in the state of the art SSM~\cite{Franceschi2020ICML}. We adapt a similar residual update strategy for predicting future BEV representations. We also experimentally show that the content variable for the static part of the scene is not as effective in the BEV space as it is in pixel space~\cite{Franceschi2020ICML}. 

\subsection{Future Prediction in Driving}
The typical approach to the prediction problem in self-driving is to first perform detection and tracking, and then do the trajectory prediction~\cite{Chai2020CORL,Hong2019CVPR}. In these methods, errors are propagated at each step. There are some recent methods~\cite{Luo2018CVPR,Casas2018CORL,Casas2020ICRA,Djuric2021IV} which directly address the prediction problem from the sensory input including LiDAR, radar, and other sensors. These methods also typically rely on an HD map of the environment. Due to high performance and efficiency of end-to-end approach, we follow a similar approach for future prediction but using cameras only and without relying on HD maps.

Despite their efficiency, most of the previous work focus on the most likely prediction~\cite{Casas2018CORL} or only models the uncertainty regarding the ego-vehicle's trajectory~\cite{Casas2020ICRA,Djuric2021IV}. The motion prediction methods which consider the behavior of all the agents in the scene typically assume a top-down rasterised representation as input, \eg Argoverse setting~\cite{Chang2019CVPR}. Even then, multiple future prediction problem is typically addressed by generating a fixed number of predictions~\cite{Gao2020CVPR,Liang2020ECCV,Aydemir2022arXiv}, for example by estimating the likelihood of multiple target locations~\cite{Zhao2020CORL,Gu2021ICCV}. There are some exceptions~\cite{Tang2019NeurIPS,Sriram2020ECCV,Huang2020RAL} which directly address the diversity aspect with a probabilistic framework. These works, especially the ones using a deep variational framework~\cite{Tang2019NeurIPS,Sriram2020ECCV} are similar to our approach in spirit, however, they operate in the coordinate space by assuming the availability of a top-down map where locations of agents are marked. We aim to learn this top-down BEV representation from multiple cameras by also segmenting the agents in the scene.

FIERY~\cite{Hu2021ICCV} is the first to address probabilistic future prediction from multiple cameras. However, future predictions are limited both in terms of diversity and length considering the typical video prediction setting. We propose a probabilistic future prediction method that can generate diverse predictions extending to different temporal horizons with a stochastic temporal dynamics model.
\section{Methodology}
\label{sec:method}
\subsection{A Compact Representation for Future Prediction}
\label{sec:bev}
Modern self-driving vehicles are typically equipped with multiple cameras observing the scene from multiple viewpoints. Placing cameras on the vehicle is cheap but processing information even from a single camera can be quite expensive. The traditional approach in computer vision is to extract low-level and semantic cues from these cameras and then fuse them into a holistic scene representation to perform prediction and planning. Recent success of end-to-end methods in driving has led to a rethinking of this approach. Furthermore, building and maintaining HD maps require a significant effort which is expensive and hard to scale. A better approach is to learn a geometrically consistent scene representation which can also mark the location, the motion, and even the semantics of the dynamic objects in the scene.

The bird's-eye view (BEV) representation initially proposed in \cite{Philion2020ECCV}, takes image $\bx_t^i$ at time $t$ from each camera $i \in \{1,\cdots,6\}$ and fuses them into a compact BEV representation $\bs_t$. This is achieved by encoding each image and also by predicting a distribution over the possible depth values. The BEV features are obtained by weighting the encoded image features according to depth probabilities predicted. These features are first lifted to 3D by using known camera intrinsic and extrinsic parameters and then the height dimension is pooled over to project the features into the bird's-eye view. This results in the state representation $\bs_t$ that we use for future prediction as explained next. 

\subsection{Learning Temporal Dynamics in BEV}
\label{sec:dynamics}
\begin{figure}[t]
\centering
    \includegraphics[width=\textwidth]{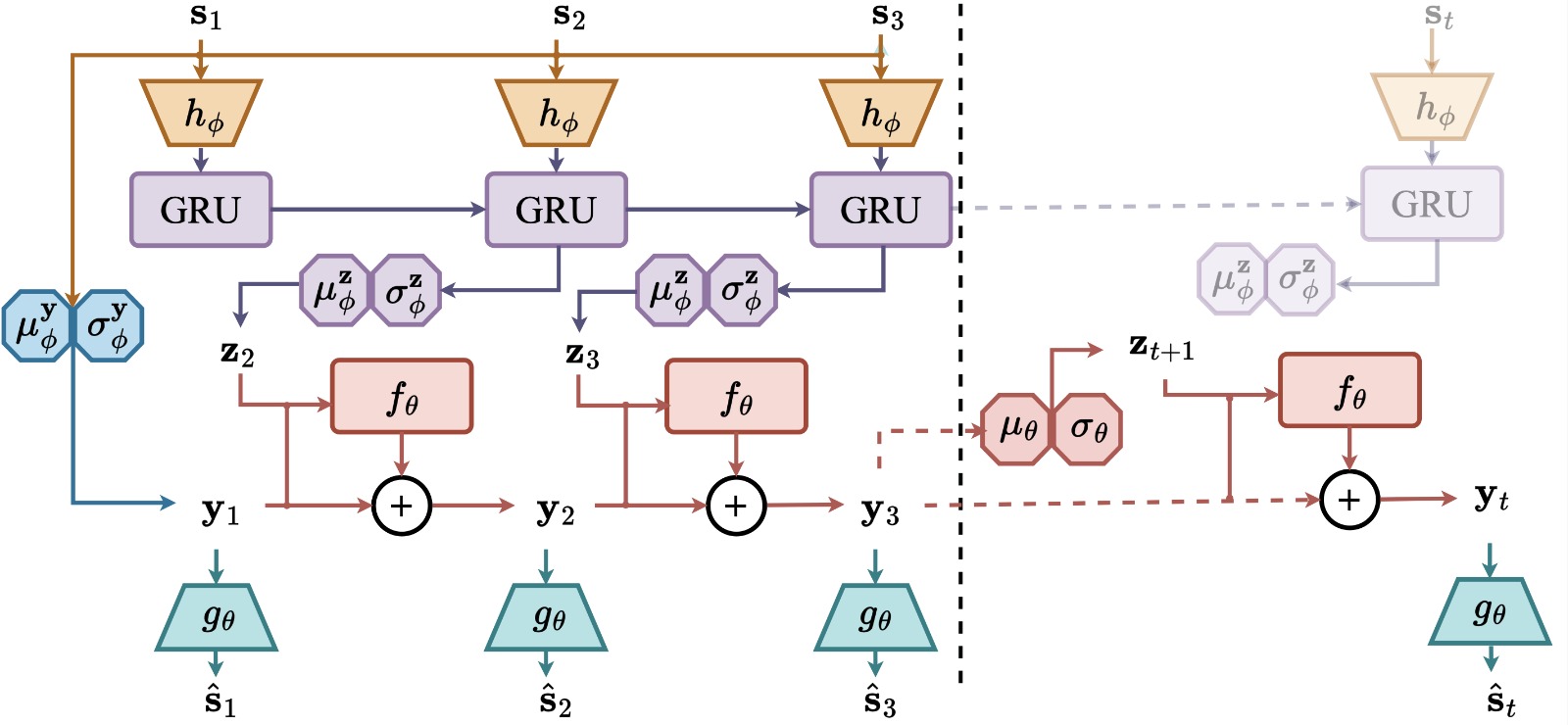}
    \caption{\textbf{Architecture for Learning Temporal Dynamics.} This figure shows the inference procedure of our model StretchBEV. We start with the first $k=3$ conditioning frames where we sample the stochastic latent variables from the posterior distribution~(purple). On the right, we show the prediction at a step $t$ after the conditioning frames where we sample from the learned future distribution~(red). The dashed vertical line marks the conditioning frames.}
    \label{fig:temporal_dynamics}
\end{figure}
\subsubsection{Notation}
In our formulation, $\bs_{1:T}$ denotes a sequence of BEV feature maps representing the state of the vehicle and its environment for $T$ time steps. In stochastic future prediction, the goal is to predict possible future states $\hat{\bs}_{k+1:T}$ conditioned on the state in the first $k$ time steps. Precisely, we condition on the first $k=3$ steps and predict future in varying lengths from 4 to 12 steps ahead. 

Differently from previous work on stochastic video prediction \cite{Denton2018ICML,Franceschi2020ICML}, the BEV state $\bs_t$ that is input to our stochastic prediction framework is the intermediate representations in a high dimensional space rather than a video frame in the pixel space as explained in \secref{sec:bev}. Similarly, the predicted output $\hat{\bs}_t$ represents the predictions of future in the same high dimensional space. We decode these high dimensional future predictions into various output modalities $\hat{\bo}_t$ such as future instance segmentation and motion as explained in \secref{sec:decoding}. While these modalities need to be trained in a supervised manner in contrast to typical self-supervised stochastic video prediction frameworks \cite{Denton2018ICML,Franceschi2020ICML}, they provide interpretability which is critical in self-driving. Furthermore, using these modalities in the posterior in addition to the future state representations improves the results significantly as we show in our experiments. 

\subsubsection{Stochastic Residual Dynamics}
Following \cite{Franceschi2020ICML}, we learn the changes in the states through time with stochastic residual updates to a sequence of latent variables. For each state $\bs_t$, there is a corresponding latent variable $\by_t$ generating it, independent of the previous states (\figref{fig:temporal_dynamics}). Each $\by_{t+1}$ only depends on the previous $\by_t$ and a stochastic variable $\bz_{t+1}$. The randomness is introduced by the stochastic latent variable $\bz_{t+1}$ which is sampled from a normal distribution learned from the previous state's latent variable only:
\begin{equation}
    \bz_{t+1} \sim \cN\left(\mu_{\theta}(\by_t), \sigma_{\theta}(\by_t)~\bI \right)
\end{equation}
Given $\bz_{t+1}$, the dependency between the latent variables $\by_t$ and $\by_{t+1}$ is deterministic through the residual update:
\begin{equation}
    \label{eq:y_res}
    \by_{t+1} = \by_t + f_{\theta}\left(\by_t, \bz_{t+1}\right)
\end{equation}
where $f_{\theta}$ is a small CNN to learn the residual updates to $\by_t$. We learn the distribution of future states from the corresponding latent variable as a normal distribution with constant diagonal variance: $\hat{\bs}_t \sim \cN(g_{\theta}(\by_t))$. The first latent variable is inferred from the conditioning frames by assuming a standard Gaussian prior: $\by_1 \sim \cN\left(\b0, \bI \right)$.

\boldparagraph{On the Content Variable}
In video prediction, a common practice is to represent the static parts of the scene with a content variable which allows the model to focus on the moving parts. On the contrary to the state of the art in video prediction~\cite{Franceschi2020ICML}, the content variable does not improve the results in our case (see Supplementary), therefore we omit it in our formulation here. This can be attributed to the details in the background that are confusing for learning dynamics while operating in the pixel space, but in our case, most of these details are already suppressed in the BEV representation. 

\boldparagraph{On Diversity}
In contrast to a present and a future distribution in FIERY \cite{Hu2020ECCV,Hu2021ICCV}, there is a distribution learned at each time step in our model. This corresponds to sampling stochastic random variables at each time step as opposed to sampling once to represent all the future frames. This is the key property which allows our model to produce diverse predictions in long sequences. By sampling from a learned distribution at each time step, our model learns to represent the complex dynamics of future frames, even for predictions further away from the conditioning frames. Furthermore, our model does not need a separate temporal block for learning the dynamics prior to learning these distributions. The dynamics are learned through the temporal evolution of latent variables by also considering the randomness of the future predictions with stochastic random variables at each time step. This not only increases the diversity of predictions but also alleviates the need for a separate temporal block, \eg with 3D convolutions. Note that our formulation is still efficient, almost the same inference time as FIERY (see Supplementary), because the latent variables are low dimensional and each state is generated independently.

Moreover, FIERY uses only a single vector of latent variables which is expanded to the spatial grid to generate futures states probabilistically. Therefore, it uses the same stochastic noise in all the coordinates of the grid. However, in our model, we have a separate random variable at each coordinate of the grid to model the uncertainty spatially as well. Training FIERY with a separate random variable at each location of the grid results in diverging loss values.

\subsection{Variational Inference and Architecture}
Following the generative process in \cite{Franceschi2020ICML}, the joint probability of the BEV states $\bs_{1:T}$, the output modalities $\bo_t$, and the latent variables $\bz_{1:T}$ and $\by_{1:T}$ is as follows: %
\begin{eqnarray}
    \label{eq:cond_prob}
    p\left(\bs_{1:T}, \bo_{1:T}, \bz_{2:T}, \by_{1:T} \right) &=& p\left(\by_1\right) \prod_{t=2}^T p\left(\bz_t,\by_t \vert \by_{t-1} \right) \prod_{t=1}^T p\left(\bo_t \vert \bs_t \right) p\left(\bs_t \vert \by_t \right) \\
    \label{eq:z_y}
    p\left(\bz_t, \by_t \vert \by_{t-1} \right) &=& p\left(\by_t \vert \by_{t-1}, \bz_t \right) p\left(\bz_t \vert \by_{t-1} \right) 
\end{eqnarray}
The relationship between $\by_t$ and $\by_{t-1}$ in $p\left(\by_t \vert \by_{t-1}, \bz_t \right)$ \eqref{eq:z_y} is deterministic through the stochastic latent residual as formulated in \eqref{eq:y_res}. Similarly for the term $p\left(\bo_t \vert \bs_t \right)$ in \eqref{eq:cond_prob}, the output modalities $\bo_t$ is learned from $\bs_t$  with a deterministic decoder in a supervised manner (\secref{sec:decoding}).

Our goal is to maximize the likelihood of the BEV states extracted from the frames (\secref{sec:bev}) and the corresponding output modalities $p\left(\bs_{1:T}, \bo_{1:T}\right)$. For that purpose, we learn a deep variational inference model $q$ parametrized by $\phi$ which is factorized as follows: 
\begin{eqnarray}
    \label{eq:fact}
    q_{Z,Y} &\triangleq& q\left(\bz_{2:T}, \by_{1:T} \vert \bs_{1:T}, \bo_{2:T} \right) \\ &=& q\left(\by_1 \vert \bs_{1:k} \right) \prod_{t=2}^T q\left(\bz_t \vert \bs_{1:t}, \bo_{2:t} \right) q\left(\by_t \vert \by_{t-1}, \bz_t \right) \nonumber
\end{eqnarray}
where $k = 3$ is the number of conditioning frames and $q\left(\by_t \vert \by_{t-1}, \bz_t \right)$ is equal to $p\left(\by_t \vert \by_{t-1}, \bz_t \right)$ with the residual update as explained above. We obtain two versions of our model by keeping (StretchBEV-P) or removing (StretchBEV) the dependency of $\bz_t$ on $\bo_{1:t}$ in the posterior in \eqref{eq:fact}.
We refer the reader to Supplementary for the derivation of the following evidence lower bound~(ELBO): 
\begin{eqnarray}
    \label{eq:elbo}
    \text{log} ~p\left( \bs_{1:T}, \bo_{1:T} \right) \ge &\cL&\left( \bs_{1:T}, \bo_{1:T}; \theta, \phi \right) \\
    \triangleq &-& D_{\text{KL}} \left( q\left( \by_1 \vert \bs_{1:k}\right) \mid\mid p\left( \by_1\right) \right)
    \nonumber \\
    &+& \nE_{\left(\tilde{\bz}_{2:T}, \tilde{\by}_{1:T}\right)\sim q_{Z,Y}}
    \Bigg[\sum_{t=1}^T \text{log}~p\left( \bs_t \vert \tilde{\by}_t\right) p\left( \bo_t \vert \bs_t\right) \nonumber \\ 
    &-& \sum_{t=2}^T D_{\text{KL}} \big( q\left( \bz_t \vert \bs_{1:t}, \bo_{1:t} \right) \mid\mid p\left( \bz_t \vert \tilde{\by}_{t-1} \right) \big)\bigg]
    \nonumber
\end{eqnarray}
where $D_{\text{KL}}$ denotes the Kullback-Leibler~(KL) divergence, $\theta$ and $\phi$ represent model and variational parameters, respectively. Following the common practice, we choose $q\left( \by_1 \vert \bs_{1:k}\right)$ and $q\left( \bz_t \vert \bs_{1:t}, \bo_{1:t} \right)$ to be factorized Gaussian for analytically computing the KL divergences and use the re-parametrization trick~\cite{Kingma2014ICLR} to compute gradients through the inferred variables. 
Then, the resulting objective function is maximizing the ELBO as defined in \eqref{eq:elbo} and minimizing the supervised losses for the output modalities (\secref{sec:decoding}).

We provide a summary of the steps in our temporal model as shown in \figref{fig:temporal_dynamics}. We start by fusing the images $\bx^i_t$ at time $t$ from each camera $i$ into the BEV state $\bs_t$ at each time step as explained in \secref{sec:bev}. 
\begin{enumerate}
    \item The resulting BEV states are still in high resolution, $\bs_t \in \nR^{C \times H \times W}$ where $(H,W) = (200,200)$. Therefore, we first process them with an encoder $h_{\phi}$ to reduce the spatial resolution to $50\times50$. 
    \item The first latent variable $\by_1$ is inferred using a convolutional neural network on the first three encoded states.
    \item  The stochastic latent variable $\bz_t$ is inferred at each time step from the respective encoded state, using a recurrent neural network which is a combination of a ConvGRU and convolutional blocks.
    \item The residual change in the dynamics is predicted with $f_{\theta}$ based on both the previous state dynamics $\by_t$ and the stochastic latent variable $\bz_{t+1}$ and added to $\by_t$ to obtain $\by_{t+1}$.
    \item From each $\by_t$, the state $\hat{\bs}_t$ is predicted in the original resolution with $g_{\theta}$.
    \item Finally, the output modalities $\hat{\bo}_t$ are decoded from the state prediction $\hat{\bs}_t$.
\end{enumerate}

\subsection{Decoding Future Predictions}
\label{sec:decoding}
Based on the predictions of the future states at each time step, we train a supervised decoder to output semantic segmentation, instance center, offsets, and future optical flow, similar to the previous work~\cite{Cheng2020CVPR,Hu2021ICCV}. The decoding function is a deterministic neural network that can be trained either jointly with the dynamics or independently, \eg later for interpretability. The output modalities show both the location and the motion of instances at each time step. The motion predicted as future flow is used to track instances. We use the same supervised loss functions for each modality as the FIERY~\cite{Hu2021ICCV}. Although the decoding of future predictions is not necessary for planning and control, for example when training a driving agent to act on predictions, these predictions provide interpretability and allow to compare the methods in terms of various metrics evaluating each modality.
\section{Experiments}
\label{sec:exp}
\subsection{Dataset and Evaluation Setting}
\label{sec:data_eval}
We evaluate the performance of the proposed approach and compare to the state of the art method, FIERY \cite{Hu2021ICCV}, on the nuScenes dataset~\cite{Caesar2020CVPR}. On the nuScenes, there are $6$ cameras with overlapping views which provide the ego vehicle with a complete view of its surroundings. The nuScenes dataset consists of $1000$ scenes with $20$ seconds long at $2$ frames per second.

We first follow the training and the evaluation setting proposed in FIERY~\cite{Hu2021ICCV} for comparison by using $1.0$ second of past context to predict $2.0$ seconds of future context. Given the sampling rate of $2$ frames per second, this setting corresponds to predicting $4$ future frames conditioned on $3$ past frames. We call this setting \emph{short} in terms of temporal length and define two more settings for longer temporal predictions. All the models are trained to predict $2.0$ seconds into the future and only the evaluation is changed to predict longer time steps. In the \emph{mid} and \emph{long} settings, we double and triple the number of future frames to predict, \ie $8$ and $12$, respectively, that corresponds to $4.0$ and $6.0$ seconds into the future. These settings are closer to the stochastic video prediction setup~\cite{Denton2018ICML,Franceschi2020ICML,Akan2021ICCV} where there are typically many more frames to predict than the conditioning frames for measuring diversity and the performance of the models further away from the conditioning frames. Note that \emph{short} and \emph{long} refer to temporal length in our evaluations as opposed spatial coverage as defined in the previous work~\cite{Hu2021ICCV}. We also evaluate in terms of spatial coverage but call it \emph{near}~($30$m$\times30$m) and \emph{far}~($100$m $\times 100$m) for clarity.

\subsection{Training Details}
Our models follow the input and output setting proposed in the previous work \cite{Hu2021ICCV}. We process 6 camera images at a resolution of $224 \times 480$ pixels for each frame and construct the BEV state of size $200 \times 200 \times 64$. We further process the states into a smaller spatial resolution ($50 \times 50$) for efficiency before learning the temporal dynamics but increase it back to the initial resolution afterwards. Given the predicted states, we use the same decoder architecture as the FIERY to decode the object centers, the segmentation masks, the instance offsets, and the future optical flow at a resolution of $200 \times 200$ pixels. We provide the details of the architecture in Supplementary and we will release the code upon publication.

In our approach, learning temporal dynamics and decoding output modalities are separated from each other. Therefore, we can pre-train the temporal dynamics part without using the labels for the output modalities. In pre-training, our objective is to learn to match the future states that are extracted using a pre-trained BEV segmentation model~\cite{Philion2020ECCV}, conditioned on the past states. This approach is more similar to self-supervised stochastic video prediction methods \cite{Denton2018ICML,Franceschi2020ICML}. Furthermore, this way, learning of temporal dynamics can be improved by using camera sequences only as input which can be easily collected in large quantities. Then, we fine-tune the temporal dynamics with a smaller learning rate (see Supplementary for details) while learning to decode the output modalities in a supervised manner. The alternative is to jointly train the temporal dynamics and supervised decoding without pre-training. We present the results of our model StretchBEV with and without pre-training.

StretchBEV does not use the labels ($\bo_t$) for learning the temporal dynamics, it only uses them in the supervised loss to decode the output modalities. In our full model StretchBEV-P, we encode output modalities following FIERY and use them in the posterior for learning the temporal dynamics. During training, we sample the stochastic latent variables from the posterior and learn to minimize the difference between the posterior and the future distribution. During inference, we sample from the posterior in the conditioning frames and sample from the learned future distribution in the following steps as shown in \figref{fig:temporal_dynamics}.

\subsection{Metrics}
We use two different metrics for evaluating the decoded modalities, one frame level and another video level, that are also used in the previous work~\cite{Hu2021ICCV}. The first is Intersection over Union~(IoU) to measure the quality of the segmentation at each frame. The second is Video Panoptic Quality~(VPQ) to measure the quality of the segmentation and consistency of the instances through the video.

We evaluate the diversity quantitatively in terms of Generalized Energy Distance ($D_\mathrm{GED}$) \cite{GED} by using $\left(1 - \mathrm{VPQ}\right)$ as the distance as proposed in FIERY~\cite{Hu2021ICCV}. 

\begin{figure}[t]
\centering
    \includegraphics[width=\textwidth]{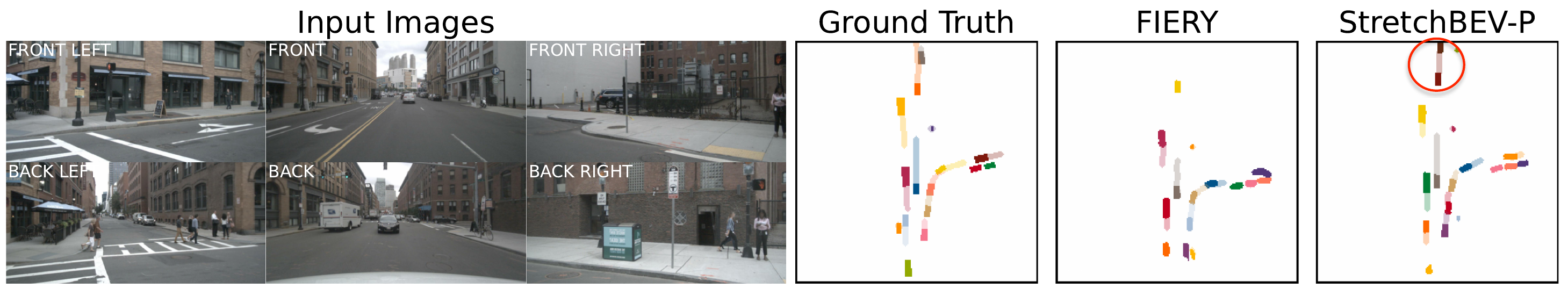} %
    \\
    \includegraphics[width=\textwidth]{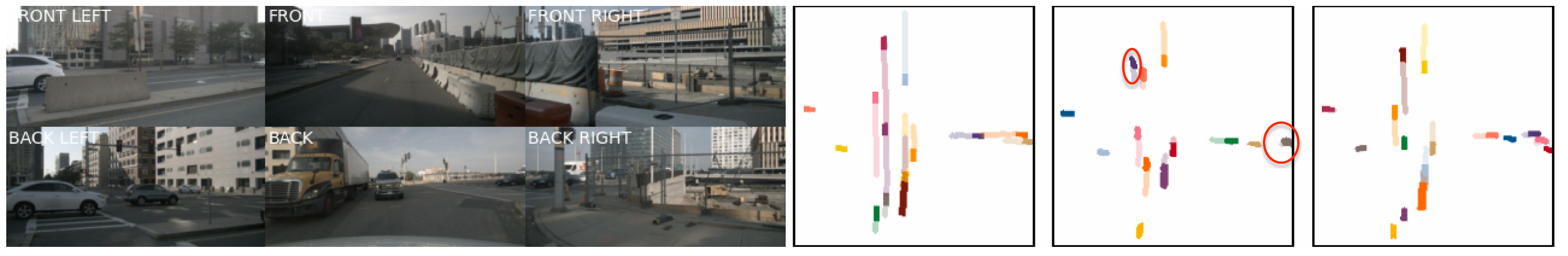} \\ %
    \includegraphics[width=\textwidth]{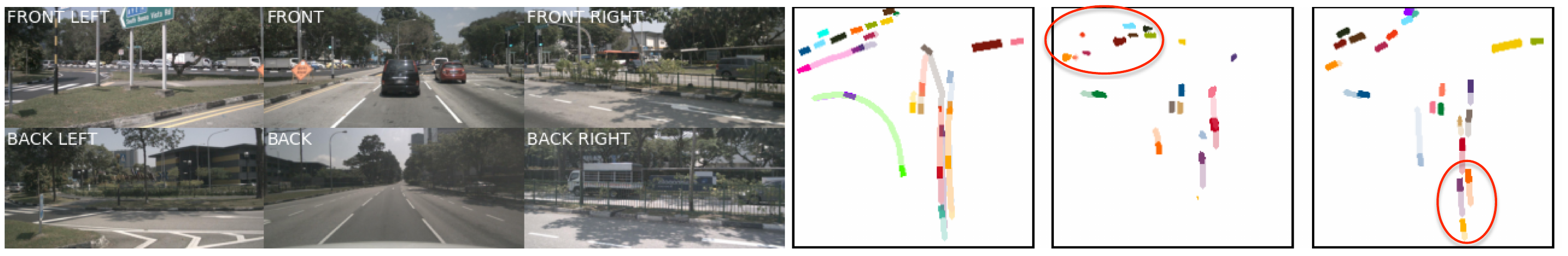} 
    \caption{\textbf{Qualitative Comparison over Different Temporal Horizons.} In this figure, we qualitatively compare the results of our model StretchBEV-P \textbf{(right)} to the ground truth \textbf{(left)} and FIERY \cite{Hu2021ICCV} \textbf{(middle)} over short \textbf{(top)}, mid \textbf{(middle)}, and long \textbf{(bottom)} temporal horizons. Each color represents an instance of a vehicle with its trajectory trailing in the same color transparently.}
    \label{fig:comp_temp}
\end{figure}

\subsection{Ablation Study}
\begin{table}[t]
    \centering
    \setlength{\tabcolsep}{7pt} 
    \begin{tabular}{c | c | c | c c | c c}
     & Pre- & ~Posterior~ & \multicolumn{2}{c|}{IoU ($\uparrow$)} & \multicolumn{2}{c}{VPQ ($\uparrow$)} \\ 
    \textbf{} & ~training~ & w/labels & ~Near~ & ~Far~ & ~Near~ & ~Far~ \\ \toprule
    \multirow{2}{*}{~StretchBEV~} & \textemdash & \multirow{2}{*}{\textemdash} & 53.3 & 35.8 & 41.7 & 26.0 \\
     & \checkmark & & 55.5 & 37.1 & 46.0 & 29.0 \\ 
    \midrule
     FIERY~\cite{Hu2021ICCV} & \multirow{2}{*}{\textemdash} & \multirow{2}{*}{\checkmark} & \textbf{59.4} & 36.7 & 50.2 & 29.4 \\
    Reproduced & & & 58.8 & 35.8 & 50.5 & 29.0 \\
    \midrule
    StretchBEV-P & \textemdash &  \checkmark & 58.1 & \textbf{52.5} & \textbf{53.0} & \textbf{47.5} \\
    \bottomrule
    \end{tabular}
    \caption{\textbf{Ablation Study.} In this table, we present the results for the two versions of our model with (StretchBEV-P) and without (StretchBEV) using the labels for the output modalities in the posterior while learning the temporal dynamics and also show the effect of pre-training for the latter in comparison to FIERY \cite{Hu2021ICCV} and our reproduced version of their results (Reproduced).}
    \label{tab:ablation}
\end{table}
In \tabref{tab:ablation}, we evaluate the effect of different versions of our model using IoU and VPQ metrics in the short temporal setting to be comparable to the previous work FIERY~\cite{Hu2021ICCV}. We reproduced their results as shown in the row \emph{Reproduced}. In the first part of the table~(StretchBEV), we show the results without explicitly using the labels for future prediction. In that case, labels are only used for decoding the output modalities and back-propagated to future prediction through decoding. Although this introduces a two-stage training, we believe that reporting results using this separation is important for future work to focus on future prediction with more unlabelled data.
We measure the effect of pre-training by learning to match our future predictions to the results of a pre-trained model~\cite{Philion2020ECCV} in terms of the BEV state representation. Pre-training allows our model to learn the dynamics before decoding and improves the results significantly in each metric.

In the second half of the \tabref{tab:ablation}, we report the results using the labels in future prediction by explicitly feeding their encoding to the posterior distribution with the same encoding used in \cite{Hu2021ICCV} to learn the future distribution. The difference between StretchBEV and StretchBEV-P is that the first has access to the BEV encoding of future predictions while the latter has access to both the BEV encoding and the encoding of the output modalities to predict in the posterior distribution. As can be seen from the results, both FIERY and our model using the labels in the future distribution perform better. This shows the importance of using a more direct and accurate information about future while learning the posterior. Compared to FIERY, our model can use the labels in the conditioning frames during inference and improves the results, especially in spatially far regions and in terms of VPQ, which point to a higher quality in our predictions stretching spatially and temporally over the video. Please see Supplementary for an extended version of \tabref{tab:ablation} with multiple samples including standard deviation as an indication of uncertainty.

\subsection{Temporally Long Predictions}
\begin{figure}[t]
\centering
    \includegraphics[width=\textwidth]{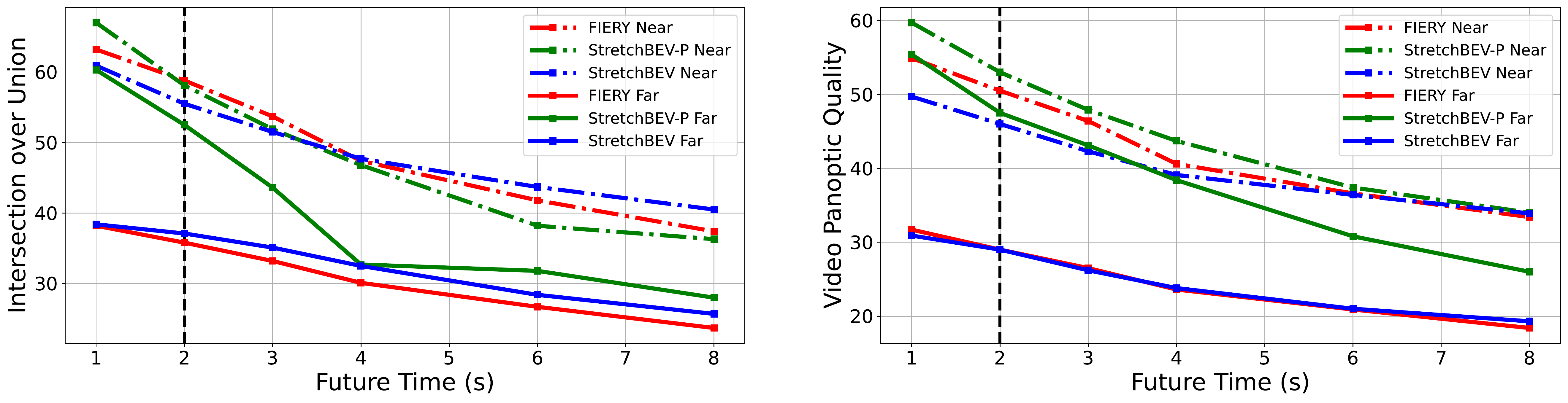}
    \caption{\textbf{Evaluation over Different Temporal Horizons.} We plot the performance of our models StretchBEV and StretchBEV-P in comparison to FIERY~\cite{Hu2021ICCV} over a range of temporal horizons from 1 second to 8 seconds in terms of IoU~\textbf{(left)} and VPQ~\textbf{(right)} for spatially far~\textbf{(solid)} and near~\textbf{(dashed)} regions separately. The vertical dashed line marks the training horizon. %
    }
    \label{fig:temporal_horizon}
\end{figure}
In longer temporal horizons, future prediction becomes increasingly difficult. This is mainly due to increasing uncertainty of future further away from conditioning frames. %
In \figref{fig:temporal_horizon},
we present the results over different temporal horizons for our model with pre-training without using the labels in the posterior (StretchBEV), FIERY~\cite{Hu2021ICCV}, and our model by using the labels in the posterior (StretchBEV-P). There is a separate plot for IoU on the left and for VPQ on the right with respect to the future time steps predicted, ranging from $1$s to $8$s. The vertical line in $2$s marks the training horizon. 
In Supplementary, we provide a table for the results over short, mid, and long temporal spans.

The negative effect of uncertain futures on each metric can be observed from the results of all the methods degrading from shorter to longer temporal spans. Our models perform better than FIERY in longer temporal spans. This is due to better handling of uncertainty with stochastic latent residual variables. Our method StretchBEV-P outperforms FIERY by significant margins, especially in terms of far VPQ in longer temporal horizons, showing consistent predictions in the overall scene throughout the video. This can be attributed to the difficulty of locating small vehicles in spatially far regions. Since StretchBEV-P has access to the labels via the posterior in the conditioning frames, it learns the temporal dynamics to correctly propagate them to the future frames, while StretchBEV and FIERY struggle to locate the instances in the first place. FIERY learns a single distribution for present and future each, therefore we cannot utilize the labels in the conditioning frames with FIERY. The results of StretchBEV outperforming the other two methods in terms of near IOU in longer temporal spans is promising for future prediction methods with less supervision. 

In \figref{fig:comp_temp}, we qualitatively compare the performance of our model StretchBEV-P on the right to FIERY in the middle over short, mid, and long temporal horizons in each row. In the first row, our model predicts the future trajectories that are more similar to the ground truth shown on the left. For example, FIERY fails to predict the trajectory of the vehicle in front (marked with red circle). In the second row, our model correctly segments the vehicles, whereas FIERY misses several vehicles far on the right and also, predicts a vehicle that does not exist (in purple on the top left). In the third row, our model predicts the future trajectories of the vehicles correctly while FIERY misses some of the vehicles (marked with red circles). The challenging case of a vehicle turning on the left (green in ground truth) is missed by both models. Some of the vehicles are not visible on the input images, \eg the back camera in the long temporal horizon. We provide a gif version of these results and more examples in Supplementary.

\subsection{Segmentation}
\begin{table}[ht]
    \centering
    \setlength{\tabcolsep}{5pt} 
    \begin{tabular}{c c c c c}
    Fishing-Cam~\cite{Hendy2020CVPRW} & Fishing-LiDAR~\cite{Hendy2020CVPRW} & FIERY~\cite{Hu2021ICCV} & StretchBEV & StretchBEV-P \\ \toprule
    30.0 & 44.3 & 57.3 & \underline{58.8} & \textbf{65.7} \\ \bottomrule
    \end{tabular}
    \caption{\textbf{Comparison of Semantic Segmentation Prediction.} In this table, we compare the predictions of our models, StretchBEV and StretchBEV-P for semantic segmentation to other BEV prediction methods in terms of IoU using the setting proposed in \cite{Hendy2020CVPRW}, \ie $32.0$m $\times$ $19.2$m at $10$cm resolution over $2$s future.}
    \label{tab:seg}
\end{table}
The previous work on bird's-eye view segmentation typically focuses on single image segmentation task with a couple of exceptions focusing on prediction. In \tabref{tab:seg}, we compare to two BEV segmentation prediction methods~\cite{Hendy2020CVPRW,Hu2021ICCV} using their setting with $32.0$m $\times$ $19.2$m at $10$cm resolution. Both methods predict $2$s into the future which corresponds to our short temporal setting. FIERY~\cite{Hu2021ICCV} outperforms the previous method~\cite{Hendy2020CVPRW} even when using LiDAR, and our method significantly outperforms both methods. %

\subsection{Diversity}
\begin{table}[b]
    \centering
    \setlength{\tabcolsep}{7pt} 
    \begin{tabular}{c | c c | c c | c c}
    \multicolumn{1}{c}{} & \multicolumn{6}{c}{Generalized Energy Distance ($\downarrow$)} \\
    \cmidrule(){2-7}
    \multicolumn{1}{c}{} & \multicolumn{2}{c}{Short} & \multicolumn{2}{c}{Mid} & \multicolumn{2}{c}{Long} \\
    \cmidrule(r){2-3} \cmidrule(r){4-5} \cmidrule(r){6-7}
    \multicolumn{1}{c|}{} &  Near & Far & Near & Far & Near & Far  \\ 
    \toprule
    FIERY~\cite{Hu2021ICCV} & 106.09 & 140.36 & 118.74 & 147.26 & 127.18 & 152.38 \\  
    StretchBEV & \underline{103.97} & \underline{132.38} & \underline{114.11} & \underline{138.15} & \underline{119.01} & \underline{142.51} \\
    StretchBEV-P &  \textbf{82.04} & \textbf{85.51} & \textbf{94.02} & \textbf{98.45} & \textbf{101.90} & \textbf{109.12} \\
    \bottomrule
    \end{tabular}
    \caption{\textbf{Quantitative Evaluation of Diversity.} This table compares the results of our models to the reproduced results of FIERY~\cite{Hu2021ICCV} in terms of Generalized Energy Distance based on VPQ~(lower better) for evaluating diversity.} %
    \label{tab:ged}
\end{table}
\begin{figure}[t]
\centering
    \includegraphics[width=.49\textwidth]{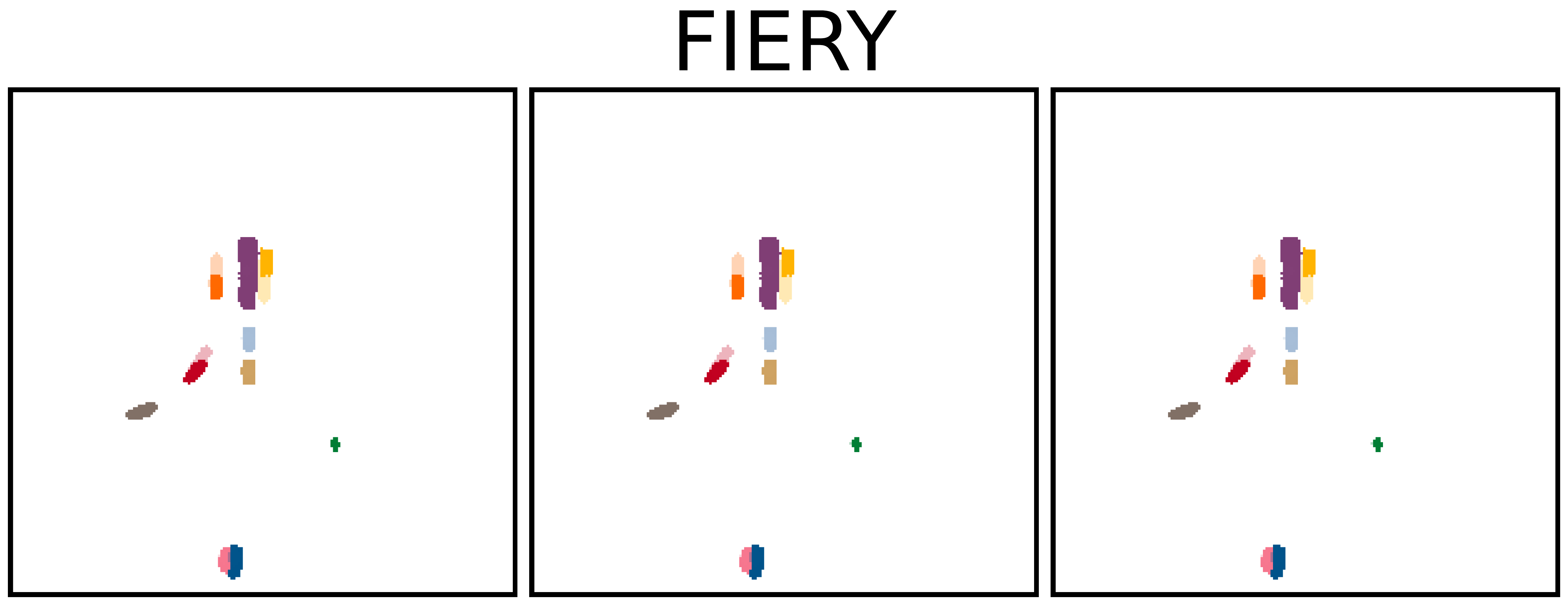}  \includegraphics[width=.49\textwidth]{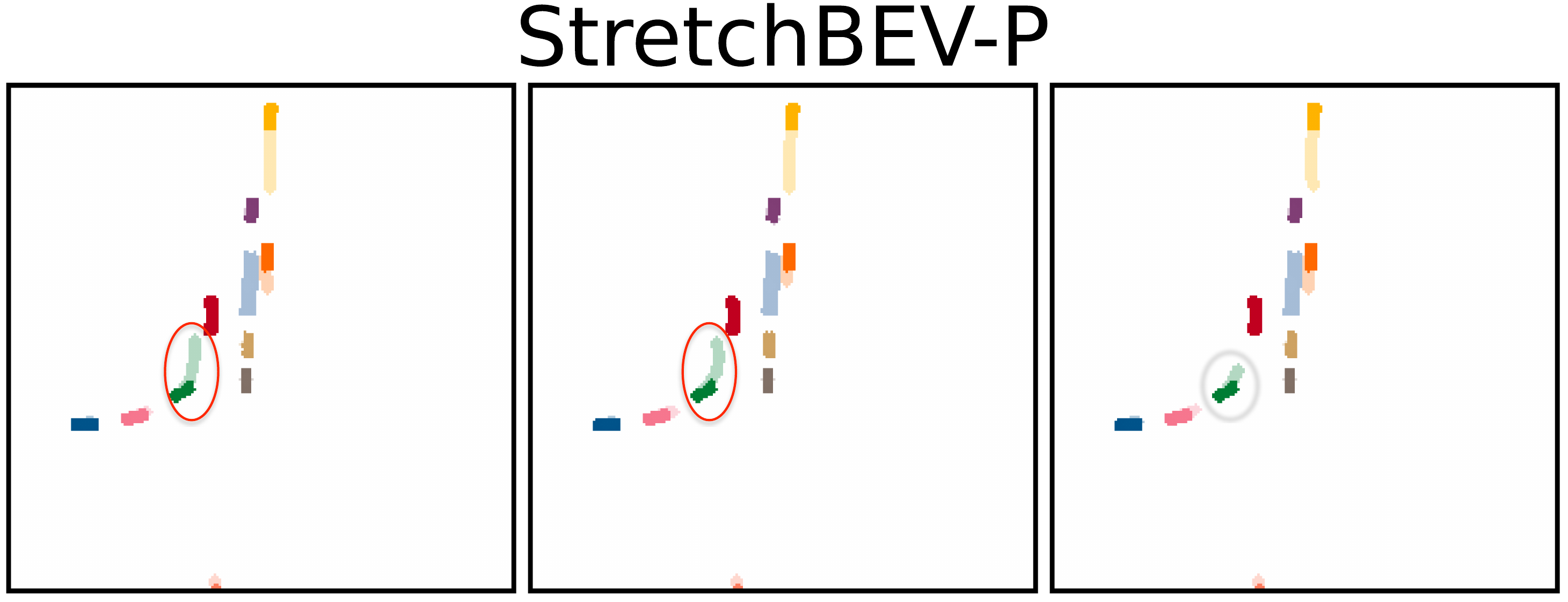} \\
    \includegraphics[width=.49\textwidth]{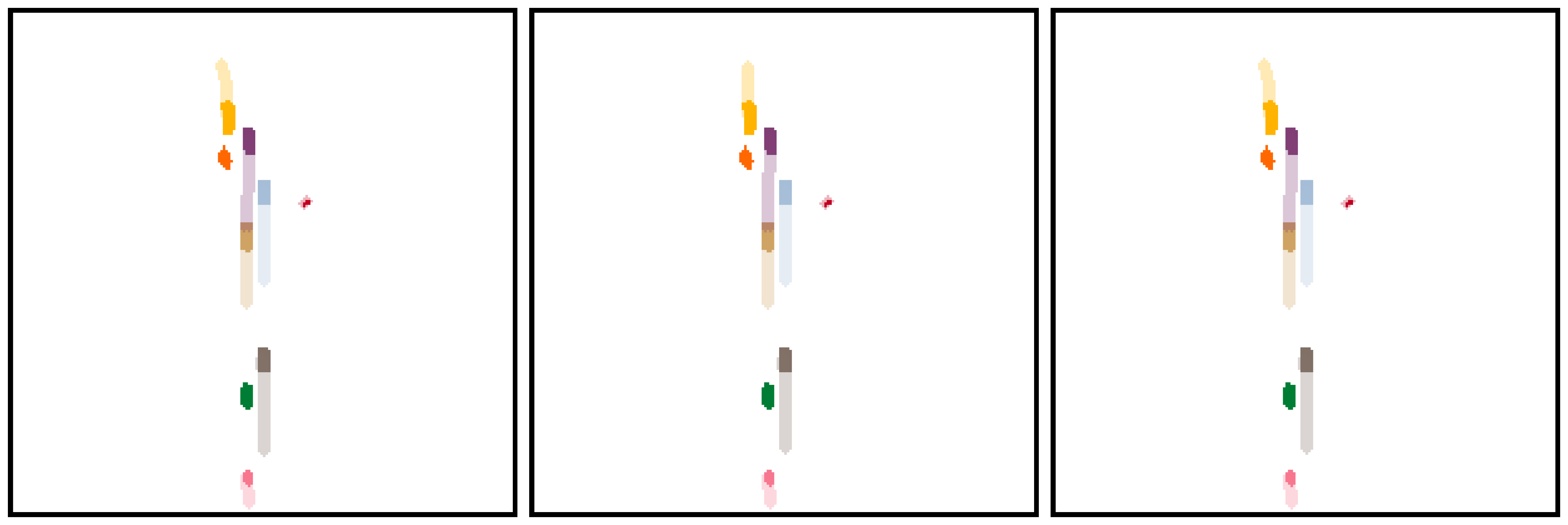} \includegraphics[width=.49\textwidth]{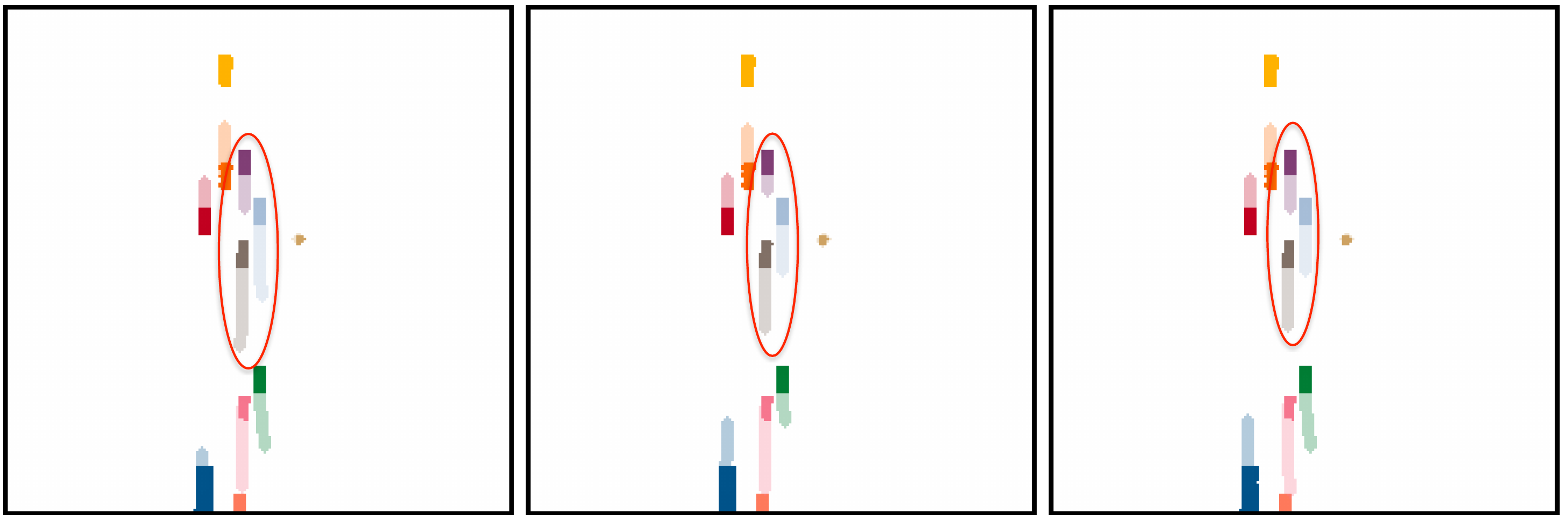} \\
    \includegraphics[width=.49\textwidth]{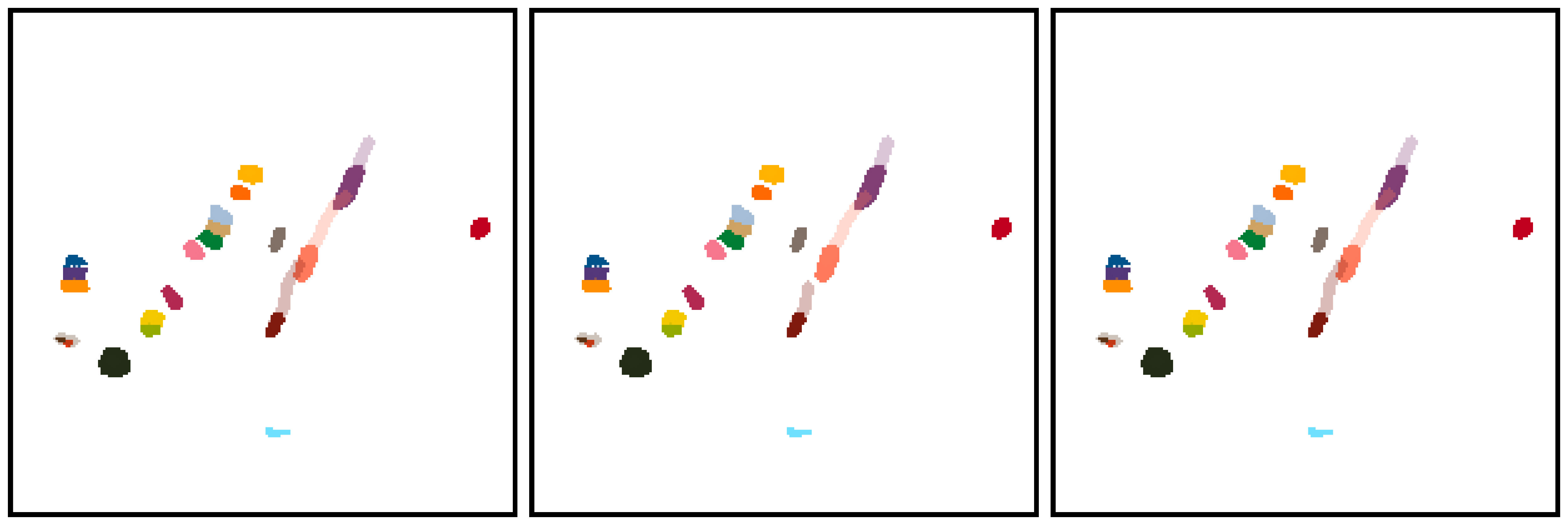} \includegraphics[width=.49\textwidth]{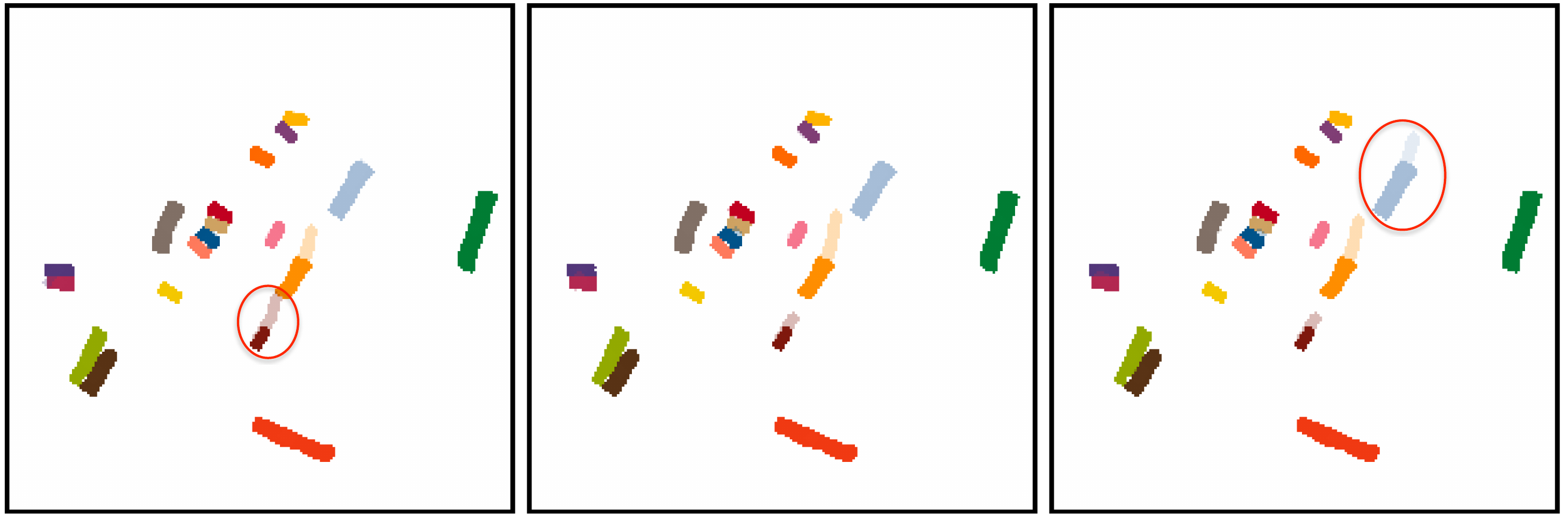}
    \caption{\textbf{Qualitative Comparison of Diversity.} In this figure, we visualize random samples from FIERY \cite{Hu2021ICCV} \textbf{(left)} and our model StretchBEV-P \textbf{(right)} over short~\textbf{(top)}, mid~\textbf{(middle)}, and long~\textbf{(bottom)} temporal horizons.}
    \label{fig:comp_samples}
\end{figure}
We quantitatively evaluate diversity by computing $D_\mathrm{GED}$ over 10 samples and show the results in \tabref{tab:ged} for our model StretchBEV-P and FIERY (our reproduced version). Our model outperforms FIERY with lower distance scores, demonstrating higher levels of diversity in the samples quantitatively. The difference is especially apparent in spatially far regions.
For qualitative comparison, in \figref{fig:comp_samples}, we visualize three samples from FIERY (left) and our model (right) over short, mid, and long temporal spans from top to bottom. While FIERY generates almost the same predictions in all three samples, our model can generate diverse predictions of future (marked with red). In the first row, our model can predict the turning behavior of the green vehicle at different speeds. In the second row, our model learns to adjust the speed of nearby vehicles proportionally, as in the case of purple, blue, and gray vehicles in the middle. Similarly, in the third row, our model can generate different predictions for the moving vehicles in the middle.
Please see Supplementary for the gif results with more examples.

\section{Conclusion and Future Work}
We introduced StretchBEV, a stochastic future instance prediction method that improve over the state of the art, especially in challenging cases, with more diverse predictions. We proposed two versions of our method with and without the labels for output modalities  explicitly in the posterior while learning the dynamics. Both models improve the state of the art in spatially far regions and over temporally long horizons. Using labels in the posterior significantly improves the results in almost all metrics but introduces a dependency on the availability of labels in the conditioning frames during inference. Future work on learning dynamics should focus on closing the gap between the two approaches, for example with scheduled sampling.

Our temporal dynamics model can be interpreted as a Neural-ODE \cite{Chen2018NeurIPS} because of its residual update dynamics. In our model, we use only one update in between time steps but in future, we plan to explore increasing the number of updates in between time step as done in the previous work~\cite{Franceschi2020ICML}.
We showed that our model increases the diversity of predictions due to improved modeling of stochasticity with sampling at every time step.
In future, we plan to explore driving policies that can utilize stochastic future predictions. Learned latent states at each time step can be directly fed into a policy learning algorithm, \eg as states in deep reinforcement learning. Furthermore, these states can be interpreted via supervised decoding into various future modalities that we predict.

\boldparagraph{Acknowledgments}
K. Akan was supported by KUIS AI Center fellowship, F. G\"{u}ney by TUBITAK 2232 International Fellowship for Outstanding Researchers.

\clearpage
\bibliographystyle{splncs04}
\bibliography{bibliography_long, ref}

\begin{thebibliography}{10}
\providecommand{\url}[1]{\texttt{#1}}
\providecommand{\urlprefix}{URL }
\providecommand{\doi}[1]{https://doi.org/#1}

\bibitem{Akan2021ICCV}
Akan, A.K., Erdem, E., Erdem, A., G\"uney, F.: Slamp: Stochastic latent
  appearance and motion prediction. In: Proc. of the IEEE International Conf.
  on Computer Vision (ICCV) (2021)

\bibitem{Aydemir2022arXiv}
Aydemir, G., Akan, A.K., G{\"u}ney, F.: Trajectory forecasting on temporal
  graphs. arXiv preprint arXiv:2207.00255  (2022)

\bibitem{Babaeizadeh2018ICLR}
Babaeizadeh, M., Finn, C., Erhan, D., Campbell, R.H., Levine, S.: Stochastic
  variational video prediction. In: Proc. of the International Conf. on
  Learning Representations (ICLR) (2018)

\bibitem{Caesar2020CVPR}
Caesar, H., Bankiti, V., Lang, A.H., Vora, S., Liong, V.E., Xu, Q., Krishnan,
  A., Pan, Y., Baldan, G., Beijbom, O.: nuscenes: A multimodal dataset for
  autonomous driving. In: Proc. IEEE Conf. on Computer Vision and Pattern
  Recognition (CVPR) (2020)

\bibitem{Casas2020ICRA}
Casas, S., Gulino, C., Liao, R., Urtasun, R.: Spagnn: Spatially-aware graph
  neural networks for relational behavior forecasting from sensor data. In:
  Proc. IEEE International Conf. on Robotics and Automation (ICRA) (2020)

\bibitem{Casas2018CORL}
Casas, S., Luo, W., Urtasun, R.: Intentnet: Learning to predict intention from
  raw sensor data. In: Proc. Conf. on Robot Learning (CoRL) (2018)

\bibitem{Castrejon2019ICCV}
Castrejon, L., Ballas, N., Courville, A.: Improved conditional vrnns for video
  prediction. In: Proc. of the IEEE International Conf. on Computer Vision
  (ICCV) (2019)

\bibitem{Chai2020CORL}
Chai, Y., Sapp, B., Bansal, M., Anguelov, D.: Multipath: Multiple probabilistic
  anchor trajectory hypotheses for behavior prediction. In: Proc. Conf. on
  Robot Learning (CoRL) (2020)

\bibitem{Chang2019CVPR}
Chang, M.F., Lambert, J.W., Sangkloy, P., Singh, J., Bak, S., Hartnett, A.,
  Wang, D., Carr, P., Lucey, S., Ramanan, D., Hays, J.: Argoverse: 3d tracking
  and forecasting with rich maps. In: Proc. IEEE Conf. on Computer Vision and
  Pattern Recognition (CVPR) (2019)

\bibitem{Chen2018NeurIPS}
Chen, R.T., Rubanova, Y., Bettencourt, J., Duvenaud, D.K.: Neural ordinary
  differential equations. In: Advances in Neural Information Processing Systems
  (NeurIPS) (2018)

\bibitem{Cheng2020CVPR}
Cheng, B., Collins, M.D., Zhu, Y., Liu, T., Huang, T.S., Adam, H., Chen, L.C.:
  Panoptic-deeplab: A simple, strong, and fast baseline for bottom-up panoptic
  segmentation. In: CVPR (2020)

\bibitem{Denton2018ICML}
Denton, E., Fergus, R.: Stochastic video generation with a learned prior. In:
  Proc. of the International Conf. on Machine learning (ICML) (2018)

\bibitem{Djuric2021IV}
Djuric, N., Cui, H., Su, Z., Wu, S., Wang, H., Chou, F., Martin, L.S., Feng,
  S., Hu, R., Xu, Y., Dayan, A., Zhang, S., Becker, B.C., Meyer, G.P.,
  Vallespi{-}Gonzalez, C., Wellington, C.K.: Multixnet: Multiclass multistage
  multimodal motion prediction. In: Proc. IEEE Intelligent Vehicles Symposium
  (IV) (2021)

\bibitem{Franceschi2020ICML}
Franceschi, J.Y., Delasalles, E., Chen, M., Lamprier, S., Gallinari, P.:
  Stochastic latent residual video prediction. In: Proc. of the International
  Conf. on Machine learning (ICML) (2020)

\bibitem{Gao2020CVPR}
Gao, J., Sun, C., Zhao, H., Shen, Y., Anguelov, D., Li, C., Schmid, C.:
  Vectornet: Encoding {HD} maps and agent dynamics from vectorized
  representation. In: Proc. IEEE Conf. on Computer Vision and Pattern
  Recognition (CVPR) (2020)

\bibitem{Gregor2019ICLR}
Gregor, K., Besse, F.: Temporal difference variational auto-encoder. In: Proc.
  of the International Conf. on Learning Representations (ICLR) (2019)

\bibitem{Gu2021ICCV}
Gu, J., Sun, C., Zhao, H.: Densetnt: End-to-end trajectory prediction from
  dense goal sets. In: Proc. of the IEEE International Conf. on Computer Vision
  (ICCV) (2021)

\bibitem{Hendy2020CVPRW}
Hendy, N., Sloan, C., Tian, F., Duan, P., Charchut, N., Xie, Y., Wang, C.,
  Philbin, J.: {FISHING} net: Future inference of semantic heatmaps in grids.
  In: Proc. IEEE Conf. on Computer Vision and Pattern Recognition (CVPR)
  Workshops (2020)

\bibitem{Hong2019CVPR}
Hong, J., Sapp, B., Philbin, J.: Rules of the road: Predicting driving behavior
  with a convolutional model of semantic interactions. In: Proc. IEEE Conf. on
  Computer Vision and Pattern Recognition (CVPR) (2019)

\bibitem{Hu2020ECCV}
Hu, A., Cotter, F., Mohan, N., Gurau, C., Kendall, A.: Probabilistic future
  prediction for video scene understanding. In: Proc. of the European Conf. on
  Computer Vision (ECCV) (2020)

\bibitem{Hu2021ICCV}
Hu, A., Murez, Z., Mohan, N., Dudas, S., Hawke, J., Badrinarayanan, V.,
  Cipolla, R., Kendall, A.: {FIERY}: Future instance segmentation in bird's-eye
  view from surround monocular cameras. In: Proc. of the IEEE International
  Conf. on Computer Vision (ICCV) (2021)

\bibitem{Hu2018CVPR}
Hu, J., Shen, L., Sun, G.: Squeeze-and-excitation networks. In: Proc. IEEE
  Conf. on Computer Vision and Pattern Recognition (CVPR) (2018)

\bibitem{Huang2020RAL}
Huang, X., McGill, S.G., DeCastro, J.A., Fletcher, L., Leonard, J.J., Williams,
  B.C., Rosman, G.: Diversitygan: Diversity-aware vehicle motion prediction via
  latent semantic sampling. IEEE Robotics and Automation Letters (RA-L)  (2020)

\bibitem{Karl2017ICLR}
Karl, M., Soelch, M., Bayer, J., van~der Smagt, P.: Deep variational bayes
  filters: Unsupervised learning of state space models from raw data. In: Proc.
  of the International Conf. on Learning Representations (ICLR) (2017)

\bibitem{Kingma2014ICLR}
Kingma, D.P., Welling, M.: Auto-encoding variational bayes. In: Proc. of the
  International Conf. on Learning Representations (ICLR) (2014)

\bibitem{Krishnan2017AAAI}
Krishnan, R.G., Shalit, U., Sontag, D.: Structured inference networks for
  nonlinear state space models. In: Proc. of the Conf. on Artificial
  Intelligence (AAAI) (2017)

\bibitem{Liang2020ECCV}
Liang, M., Yang, B., Hu, R., Chen, Y., Liao, R., Feng, S., Urtasun, R.:
  Learning lane graph representations for motion forecasting. In: Proc. of the
  European Conf. on Computer Vision (ECCV) (2020)

\bibitem{Luo2018CVPR}
Luo, W., Yang, B., Urtasun, R.: Fast and furious: Real time end-to-end {3D}
  detection, tracking and motion forecasting with a single convolutional net.
  In: Proc. IEEE Conf. on Computer Vision and Pattern Recognition (CVPR) (2018)

\bibitem{Minderer2019NeurIPS}
Minderer, M., Sun, C., Villegas, R., Cole, F., Murphy, K.P., Lee, H.:
  Unsupervised learning of object structure and dynamics from videos. In:
  Advances in Neural Information Processing Systems (NeurIPS) (2019)

\bibitem{Murphy2023Prob}
Murphy, K.P.: Probabilistic Machine Learning: Advanced Topics. MIT Press
  (2023), \url{probml.ai}

\bibitem{Philion2020ECCV}
Philion, J., Fidler, S.: Lift, splat, shoot: Encoding images from arbitrary
  camera rigs by implicitly unprojecting to 3d. In: Proc. of the European Conf.
  on Computer Vision (ECCV) (2020)

\bibitem{Rubanova2019NeurIPS}
Rubanova, Y., Chen, R.T.Q., Duvenaud, D.K.: Latent ordinary differential
  equations for irregularly-sampled time series. In: Advances in Neural
  Information Processing Systems (NeurIPS) (2019)

\bibitem{Sriram2020ECCV}
Sriram, N.N., Liu, B., Pittaluga, F., Chandraker, M.: {SMART}: Simultaneous
  multi-agent recurrent trajectory prediction. In: Proc. of the European Conf.
  on Computer Vision (ECCV) (2020)

\bibitem{GED}
Székely, G.J., Rizzo, M.L.: The energy of data. Annual Review of Statistics
  and Its Application  \textbf{4}(1),  447--479 (2017)

\bibitem{Tang2019NeurIPS}
Tang, Y.C., Salakhutdinov, R.: Multiple futures prediction. In: Advances in
  Neural Information Processing Systems (NeurIPS) (2019)

\bibitem{Villegas2019NeurIPS}
Villegas, R., Pathak, A., Kannan, H., Erhan, D., Le, Q.V., Lee, H.: High
  fidelity video prediction with large stochastic recurrent neural networks.
  In: Advances in Neural Information Processing Systems (NeurIPS) (2019)

\bibitem{Zhao2020CORL}
Zhao, H., Gao, J., Lan, T., Sun, C., Sapp, B., Varadarajan, B., Shen, Y., Shen,
  Y., Chai, Y., Schmid, C., Li, C., Anguelov, D.: {TNT:} target-driven
  trajectory prediction. In: Proc. Conf. on Robot Learning (CoRL) (2020)

\end{thebibliography}

\clearpage
\appendix
\begin{abstract}
In this part, we provide additional illustrations, derivations, and results for our paper ``StretchBEV: Stretching Future Instance Prediction Spatially and Temporally". We first show the full derivation of the Evidence Lower Bound (ELBO) in \secref{app:ELBO}. In \secref{sec:model_training}, we explain the architectural choices and training details. We present the detailed versions of the quantitative results in the main paper. In addition, we present more ablation experiments with the content variable and perform a comparison in terms of inference speed. We provide more qualitative results comparing our method to ground truth and FIERY~\cite{Hu2021ICCV}, and also visualizations of samples for diversity. Video visualizations are available at our website.

\end{abstract}
\section{Evidence Lower Bound}
\label{app:ELBO}

In this section, we derive the variational lower bound for the proposed model following \cite{Franceschi2020ICML}. The changes in our derivation are mainly due to excluding the content variable and including the output modalities in the derivations.

Using the original variational lower bound of variational autoencoders \cite{Kingma2014ICLR} in \eqref{eq:OriginalELBO}:
{\setlength{\mathindent}{-0.3cm}
\begingroup
\allowdisplaybreaks
\begin{align}
    \nonumber  \text{log} ~p ( & \bs_{1:T}, \bo_{1:T} )  \\
    \label{eq:OriginalELBO}
    \geq {} &  \nE_{\left(\tilde{\bz}_{2:T}, \tilde{\by}_{1:T}\right)\sim q_{Z,Y}} \text{log}~p (\bs_{1:T}, \bo_{1:T} \vert \tilde{\bz}_{2:T}, \tilde{\by}_{1:T}) - D_{\text{KL}} (q_{Z,Y} \mid\mid p(\by_{1:T}, \bz_{2:T})) \\
    \label{eq:ModelDependencies}
    = {} &  \nE_{\left(\tilde{\bz}_{2:T}, \tilde{\by}_{1:T}\right)\sim q_{Z,Y}} \text{log}~p \left(\bs_{1:T}, \bo_{1:T}\vert \tilde{\bz}_{2:T}, \tilde{\by}_{1:T}\right)  \\
    & \hspace{3cm} - D_{\text{KL}} (q\left(\by_{1}, \bz_{2:T} \vert \bs_{1:T}, \bo_{1:T}\right) \mid\mid p(\by_{1:T}, \bz_{2:T})) \nonumber \\
    \label{eq:MutualIndependenceX}
    = {} & \nE_{\left(\tilde{\bz}_{2:T}, \tilde{\by}_{1:T}\right)\sim q_{Z,Y}} \sum\limits_{t=1}^{T} \text{log}~p(\bs_t \vert \tilde{\by}_{t}) + \text{log}~p(\bo_t \vert \bs_t) \\
    & \hspace{3cm} - D_{\text{KL}} (q(\by_{1}, \bz_{2:T} \vert \bs_{1:T}, \bo_{1:T}) \mid\mid p(\by_{1:T}, \bz_{2:T})) \nonumber
\end{align}
where:
\begin{itemize}
    \item \eqref{eq:ModelDependencies} is given by the forward and inference models factorizing $p$ and $q$ in Equations (3,4,5) in the main paper.
        \item The $\by_{2:T}$ variables are deterministic functions of $\by_{1}$ and $\bz_{2:T}$ with respect to $p$ and $q$;
    \item \eqref{eq:MutualIndependenceX} results from the factorization of $p(\bs_{1:T} \vert \by_{1:T}, \bz_{1:T})$ in Equation (3) in the main paper.
    \item $\text{log}~p(\bo_t \vert \bs_t)$ is also deterministic and corresponds to supervised decoding of output modalities (Sec 3.4 in the main paper).
\end{itemize}
From there, by using the integral formulation of $D_{\text{KL}}$:

\begin{align}
\nonumber & \text{log} ~p (  \bs_{1:T}, \bo_{1:T} )   \\
\begin{split}
        & \ge {}  \nE_{\left(\tilde{\bz}_{2:T}, \tilde{\by}_{1:T}\right)\sim q_{Z,Y}} \sum\limits_{t=1}^{T} \text{log}~p(\bs_t,  \bo_t \vert \tilde{\by}_{t}) \\
        & + \idotsint_{\by_{1}, \bz_{2:T}} q(\by_{1}, \bz_{2:T} \vert \bs_{1:T}, \bo_{1:T}) \text{log} \frac{p(\by_{1}, \bz_{2:T})}{q(\by_{1}, \bz_{2:T} \vert \bs_{1:T}, \bo_{1:T})} \text{d} \bz_{2:T} \text{d} \by_{1}
\end{split} \\
\begin{split}
        & = {}  \nE_{\left(\tilde{\bz}_{2:T}, \tilde{\by}_{1:T}\right)\sim q_{Z,Y}} \sum\limits_{t=1}^{T} \text{log}~p(\bs_t \vert \tilde{\by}_{t}) + \text{log}~p(\bo_t \vert \bs_t) - D_{\text{KL}} (q(\by_1 \vert \bs_{1:T}) \mid\mid p(\by_1))\\
        & + \nE_{\tilde{\by}_{1} \sim q(\by_1 \vert \bs_{1:T})} \bigg[\idotsint_{\bz_{2:T}} q(\bz_{2:T} \vert \bs_{1:T}, \bo_{1:T}, \tilde{\by}_1) \text{log} \frac{p(\bz_{2:T} \vert \tilde{\by}_1)}{q(\bz_{2:T} \vert \bs_{1:T}, \bo_{1:T}, \tilde{\by}_1)} \text{d}\bz_{2:T}\bigg]
    \end{split} \\
\begin{split}
\label{eq:ModelDependencies2}
        & = {}  \nE_{\left(\tilde{\bz}_{2:T}, \tilde{\by}_{1:T}\right)\sim q_{Z,Y}} \sum\limits_{t=1}^{T} \text{log}~p(\bs_t \vert \tilde{\by}_{t}) + \text{log}~p(\bo_t \vert \bs_t) - D_{\text{KL}} (q(\by_1 \vert \bs_{1:k}) \mid\mid p(\by_1))\\
        & + \nE_{\tilde{\by}_{1} \sim q(\by_1 \vert \bs_{1:k})} \bigg[\idotsint_{\bz_{2:T}} q(\bz_{2:T} \vert \bs_{1:T}, \bo_{1:T}, \tilde{\by_1}) \text{log} \frac{p(\bz_{2:T} \vert \tilde{\by}_1)}{q(\bz_{2:T} \vert \bs_{1:T}, \bo_{1:T}, \tilde{\by}_1)} \text{d}\bz_{2:T}\bigg]
    \end{split} \\
\begin{split}
\label{eq:TemporalFactorization}
        & = {}  \nE_{\left(\tilde{\bz}_{2:T}, \tilde{\by}_{1:T}\right)\sim q_{Z,Y}} \sum\limits_{t=1}^{T} \text{log}~p(\bs_t \vert \tilde{\by}_{t}) + \text{log}~p(\bo_t \vert \bs_t) - D_{\text{KL}} (q(\by_1 \vert \bs_{1:k}) \mid\mid p(\by_1))\\
        & + \nE_{\tilde{\by}_{1} \sim q(\by_1 \vert \bs_{1:k})} \bigg[\idotsint_{\bz_{2:T}} \prod_{t=2}^{T} q(\bz_t \vert \bs_{1:t}, \bo_{1:t}) \sum_{t=2}^{T} \text{log} \frac{p(\bz_t \vert \tilde{\by}_1, \bz_{2:t-1})}{q(\bz_t \vert \bs_{1:t}, \bo_{1:t})} \text{d}\bz_{2:T}\bigg]
    \end{split}\\
\begin{split}
        \label{eq:FirstIterationELBO}
        & = {}  \nE_{\left(\tilde{\bz}_{2:T}, \tilde{\by}_{1:T}\right)\sim q_{Z,Y}} \sum\limits_{t=1}^{T} \text{log}~p(\bs_t \vert \tilde{\by}_{t}) + \text{log}~p(\bo_t \vert \bs_t) - D_{\text{KL}} (q(\by_1 \vert \bs_{1:k}) \mid\mid p(\by_1)) \\
        & - \nE_{\tilde{\by}_{1} \sim q(\by_1 \vert \bs_{1:k})} D_{\text{KL}} (q(\bz_2 \vert \bs_{1:t}, \bo_{1:t}) \mid\mid p(\bz_2 \vert \tilde{\by}_1))\\
        & + \nE_{\tilde{\by}_{1} \sim q(\by_1 \vert \bs_{1:k})} \nE_{\tilde{\bz}_{2} \sim q(\bz_2 \vert \bs_{1:2}, \bo_{1:2})} \\ 
        & ~~~~~~~~~~~~~~~~~~~~~~~\bigg[\idotsint\limits_{\bz_{3:T}} \prod_{t=3}^{T} q(\bz_t \vert \bs_{1:t}, \bo_{1:t}) \sum_{t=3}^{T} \text{log} \frac{p(\bz_t \vert \tilde{\by}_1, \bz_{2:t-1})}{q(\bz_t \vert \bs_{1:t}, \bo_{1:t})} \text{d}\bz_{3:T}\bigg]
\end{split}
\end{align}
where:\\
\begin{itemize}
    \item \eqref{eq:ModelDependencies2} follows from the inference model of Equation (5) in the main paper, where $\by_{1}$ only depends on $\bs_{1:k}$;
    \item \eqref{eq:TemporalFactorization} is obtained from the factorizations of Equations (3,4,5) in the main paper.
\end{itemize}
By iterating \eqref{eq:FirstIterationELBO}'s step on $\bz_{3}, \ldots, \bz_{T}$ and factorizing all expectations, we obtain:
\begin{eqnarray}
\text{log}&~& p(  \bs_{1:T}, \bo_{1:T} ) \\
\geq &~& \nE_{\left(\tilde{\bz}_{2:T}, \tilde{\by}_{1:T}\right)\sim q_{Z,Y}} \sum\limits_{t=1}^{T} \text{log}~p(\bs_t \vert \tilde{\by}_{t}) + \text{log}~p(\bo_t \vert \bs_t) - D_{\text{KL}} (q(\by_1 \vert \bs_{1:k}) \mid\mid p(\by_1)) \nonumber \\
- &~& \nE_{\tilde{\by}_{1} \sim q(\by_1 \vert \bs_{1:k})} \bigg( \nE_{\tilde{\bz}_{t} \sim q(\bz_t \vert \bs_{1:t}, \bo_{1:t})}\bigg)_{t=2}^T \sum_{t=2}^T D_{\text{KL}} (q (\bz_t \vert \bs_{1:t}, \bo_{1:t}) \mid\mid p(\bz_t \vert \tilde{y}_1, \tilde{\bz}_{1:t-1})) \nonumber
\end{eqnarray}
and we finally retrieve Evidence Lower Bound in (6) in the main paper by using the factorization in (5) in the main paper:

\begin{eqnarray}
\text{log}&~& p(  \bs_{1:T}, \bo_{1:T} ) \\
\geq &~& \nE_{\left(\tilde{\bz}_{2:T}, \tilde{\by}_{1:T}\right)\sim q_{Z,Y}} \sum\limits_{t=1}^{T} \text{log}~p(\bs_t \vert \tilde{\by}_{t}) + \text{log}~p(\bo_t \vert \bs_t) - D_{\text{KL}} (q(\by_1 \vert \bs_{1:k}) \mid\mid p(\by_1)) \nonumber \\
- &~&  \nE_{\left(\tilde{\bz}_{2:T}, \tilde{\by}_{1:T}\right) \sim q_{Z,Y}}
\sum_{t=2}^T D_{\text{KL}} (q (\bz_t \vert \bs_{1:t}, \bo_{1:t}) \mid\mid p(\bz_t \vert \tilde{\by}_{t-1})) \nonumber
\end{eqnarray}

\endgroup
}
\section{Model and Training Details}
\label{sec:model_training}
In this section, we provide the details of the architectures used (\secref{sec:model_details}), and the details of the training including the hyper-parameters used in the optimization (\secref{sec:training_details}).

\subsection{Model Details}
\label{sec:model_details}
Our models use the same framework as FIERY~\cite{Hu2021ICCV} following the same input-output setting to be comparable. Both models process $n=6$ camera images at $(H_{\mathrm{in}}, W_{\mathrm{in}}) = (224\times480)$ for k conditioning time steps, \ie $k=3$, which results in $18$ images in total. The minimum depth value we consider is $D_{\text{min}} = 2.0\mathrm{m}$, which corresponds to the spatial extent of the ego-car. The maximum depth value is $D_{\text{max}}=50.0\mathrm{m}$, and the size of each depth slice is set to $D_{\mathrm{size}}=1.0\mathrm{m}$.
Our model uses the same bird’s-eye view (BEV) encoder and future instance segmentation and motion decoder as FIERY~\cite{Hu2021ICCV}. For further details, we direct reviewers into their appendix section. Next, we explain the details of each block in our model and for the missing or unclear parts, the code is attached with the submission. We will also share the code and the trained models upon publication.

\boldparagraph{Dow-sampling Encoder and Up-sampling Decoder} Our model uses another encoder-decoder pair to reduce spatial size of feature extracted by the BEV encoder. Down-sampling encoder contains $10$ convolutional layers followed by batch normalization and Leaky ReLU activation. After $2$ convolutional layers, we apply a dropout with probability of $0.25$. At the end, we apply another convolutional layer with a batch normalization but with $tanh$ activation at the end. Down-sampling encoder uses max-pooling after the second and fourth convolutions to reduce the spatial size to $1/4$th resolution.
Up-sampling decoder is the symmetric version of the down-sampling decoder. We use the ``nearest" mode up-sampling instead of the max-pooling to increase the spatial size.

\boldparagraph{The First Latent State} We encode the conditioning frames with a small CNN to learn the first latent state $\by_1$. The network contains 4 convolutions followed by batch normalization and Leaky ReLU activation. We also add a Squeeze and Excitation layer after the second and the fourth convolution to enhance the learned features. At the end, we apply a convolutional layer which outputs $\bmu_\phi^{\by}$ and $\bsigma_\phi^{\by}$, and then we use them to sample the first state, $\by_1$.

\boldparagraph{Prior Distribution} We use another CNN to learn a prior distribution from the previous latent state $\by_{t-1}$. The network is the same as the first latent state network except for the input, we feed the previous latent variable, $\by_{t-1}$ at time $t$ and it produces $\bmu_\theta$ and $\bsigma_\theta$.

\boldparagraph{Posterior Distribution} For posterior distribution, we use recurrent neural network, GRU-Conv, which is a combination of SpatialGRUs and convolutions. Our goal is to learn a posterior distribution, $\bmu_\phi^{\bz}$ and $\bsigma_\phi^{\bz}$, representing the temporal dynamics.  We first process image features extracted by the BEV encoder with our GRU-Conv network. Then, for each time step, we use the same network as the prior distribution to sample a posterior distribution. GRU-Conv contains $2$ SpatialGRUs followed by $2$ convolutional blocks, each of which contains $2$ convolutions with $1 \times 1$ and $3 \times 3$ kernel sizes.

\boldparagraph{Dynamics Update} We use a network to update intermediate latent variables $\by_t$. We feed the previous latent variable $\by_{t-1}$ and the corresponding stochastic variable $\bz_t$ at time $t$, and the output of the network is added to the previous latent variable, $\by_{t-1}$. The architecture is the same as the prior distribution architecture except that it only inputs one set of parameters at the end instead of two.

\subsection{Training Details}
\label{sec:training_details}
We train our models with $2$ V100 GPUs for $25$ epochs at most. We will release all the scripts used for training and the checkpoints of the models used for evaluation.

\boldparagraph{Pre-training} Our models can be pre-trained to learn the dynamics update in an unsupervised manner. We simply initialize the BEV encoder from a checkpoint trained to segment the present time objects in bird's-eye view \cite{Philion2020ECCV}. Moreover, we remove the future instance segmentation and the motion decoder. In this setting, our dynamics model learns to predict the BEV features of future time steps that are extracted by the BEV encoder conditioned on the features of the previous time steps. This way, our model learns to predict the future in the feature space without ground truth segmentation or motion. According to our results, the pre-training improves the results significantly for StretchBEV. We cannot do unsupervised pre-training for StretchBEV-P because it needs the ground truth labels in the posterior distribution.

\boldparagraph{Training Parameters} We train all our models for $25$ epochs at most. We use a held-out validation set for model selection. We use the maximum batch size that fits into V100 GPUs, which is $2$ for training and $12$ for pre-training. We use $3\times 10^{-4}$ as a starting learning rate. We apply learning rate decay if the validation loss does not decrease for some threshold, which is the reason for a varying number of epochs depending on the model.
\section{Detailed Quantitative Results}
In this section, we provide extended versions of the tables in the main paper.

\begin{sidewaystable}[t]
    \centering
    \begin{tabular}{c | c | c | c c | c c}
     & Pre- & ~Posterior~ & \multicolumn{2}{c|}{IoU ($\uparrow$)} & \multicolumn{2}{c}{VPQ ($\uparrow$)} \\ 
    \textBF{} & ~training~ & w/labels & ~Near~ & ~Far~ & ~Near~ & ~Far~ \\ \toprule
    \multirow{2}{*}{~StretchBEV~} & \textemdash & \multirow{2}{*}{\textemdash} & 54.0 (53.3) $\pm$ 0.40 & 36.2 (35.8 $\pm$ 0.3) & 43.6 (41.7 $\pm$ 1.1) & 27.4 (26.0 $\pm$ 0.8) \\ 
     & \checkmark & & 56.2 (55.5 $\pm$ 0.4) & 37.6 (37.1 $\pm$ 0.3) & 47.9 (46.2 $\pm$ 1.0) & 30.3 (29.0 $\pm$ 0.7) \\ 
     StretchBEV-Det & \checkmark &  \textemdash & 48.8 (48.8 $\pm$ 0) & 32.1 (32.1 $\pm$ 0) & 39.6 (39.6 $\pm$ 0) & 23.6 (23.6 $\pm$ 0) \\
    StretchBEV-MCD & \checkmark &  \textemdash & 50.3 (48.2 $\pm$ 1.25) & 32.8 (31.6 $\pm$ 0.7) & 42.4 (39.0 $\pm$ 1.9) & 25.3 (23.3 $\pm$ 1.2) \\
    \midrule
     FIERY~\cite{Hu2021ICCV} & \multirow{2}{*}{\textemdash} & \multirow{2}{*}{\checkmark} & \textBF{59.4} & 36.7 & 50.2 & 29.4 \\
    Reproduced & & & 59.0 (\textBF{58.8} $\pm$ 0.2) & 35.9 (35.8 $\pm$ 0.1) & 51.2 (50.5 $\pm$ 0.4) & 29.5 (29.0 $\pm$ 0.3) \\
    \midrule
    StretchBEV-P & \textemdash &  \checkmark & \textBF{60.6} (58.2 $\pm$ \textBF{1.4)} & \textBF{54.0 (52.5 $\pm$ 0.9)} & \textBF{56.6  ( 52.4 $\pm$ 2.4)} & \textBF{51.0 (48.3 $\pm$ 1.6)} \\
    \bottomrule
    \end{tabular}
    \caption{\textbf{Ablation Study.} In this table, we present the results for the two versions of our model with (StretchBEV-P) and without (StretchBEV) using the labels for the output modalities in the posterior while learning the temporal dynamics and also show the effect of pre-training for the latter in comparison to FIERY \cite{Hu2021ICCV} and our reproduced version of their results (Reproduced).}
    \label{tab:ablation2}
\end{sidewaystable}

\begin{table}[t]
    \centering
    \setlength{\tabcolsep}{5pt} 
    \begin{tabular}{c | c | c | c | c c | c c}
     & Pre- & ~Posterior~ & ~Content~ & \multicolumn{2}{c|}{IoU ($\uparrow$)} & \multicolumn{2}{c}{VPQ ($\uparrow$)} \\ 
    \textbf{} & ~training~  & w/labels & \textbf{} & ~Near~ & ~Far~ & ~Near~ & ~Far~ \\ \toprule
    \multirow{3}{*}{~StretchBEV~} & \textemdash & \multirow{3}{*}{\textemdash} & \textemdash& 53.3 & 35.8 & 41.7 & 26.0 \\
     &  \textemdash & & \checkmark & 51.9 & 34.1 & 40.8 & 25 \\ 
     & \checkmark & & \textemdash& 55.5 & 37.1 & 46.0 & 29.0 \\ 
    \midrule
     FIERY~\cite{Hu2021ICCV} & \multirow{2}{*}{\textemdash} & \multirow{2}{*}{\checkmark} & \multirow{2}{*}{\textemdash} & \textbf{59.4} & 36.7 & 50.2 & 29.4 \\
    Reproduced & & & & 58.8 & 35.8 & 50.5 & 29.0 \\
    \midrule
    StretchBEV-P & \textemdash &  \checkmark & \textemdash& 58.1 & \textbf{52.5} & \textbf{53.0} & \textbf{47.5} \\
    StretchBEV-P & \textemdash &  \checkmark & \checkmark & 57.6 & 51.9 & 51.5 & 46.8 \\
    \bottomrule
    \end{tabular}
    
    \caption{\textbf{Ablation Study.} Different than the table that we provide in the main paper, this table includes results with Content variable. As we stated before, content variable does not improve the results because most of the details are suppressed in the BEV representation.}
    \label{tab:content}
\end{table}
In \tabref{tab:content}, we provide our ablation table by also adding the results with the content variable. As explained in the main paper, content variable does not improve the performance of our models in contrast to the state of the art video prediction \cite{Franceschi2020ICML}, therefore omitted in our formulation.

In \tabref{tab:seg}, we provide the results of the future segmentation performances in the FISHING setting as proposed in \cite{Hendy2020CVPRW}. Differently from the main paper, this table includes the result of StretchBEV as well. Both StretchBEV and StretchBEV-P outperform FIERY~\cite{Hu2021ICCV} in this setting.

\begin{table}[t]
    \centering
    \setlength{\tabcolsep}{2.25pt} 
    \begin{tabular}{c | c c | c c | c c| c c | c c | c c}
    \multicolumn{1}{c}{} & \multicolumn{4}{c}{Short} & \multicolumn{4}{c}{Mid} & \multicolumn{4}{c}{Long} \\
    \cmidrule(r){2-5} \cmidrule(r){6-9} \cmidrule(r){10-13}
    \multicolumn{1}{c}{} & \multicolumn{2}{c}{IoU ($\uparrow$)} & \multicolumn{2}{c}{VPQ ($\uparrow$)} & \multicolumn{2}{c}{IoU ($\uparrow$)} & \multicolumn{2}{c}{VPQ ($\uparrow$)} & \multicolumn{2}{c}{IoU ($\uparrow$)} & \multicolumn{2}{c}{VPQ ($\uparrow$)} \\  
    \cmidrule(r){2-3} \cmidrule(r){4-5} \cmidrule(r){6-7} \cmidrule(r){8-9} \cmidrule(r){10-11} \cmidrule(r){12-13}
    \multicolumn{1}{c|}{} &  Near & Far & Near & Far & Near & Far & Near & Far & Near & Far & Near & Far \\ 
    \toprule
    StretchBEV &  55.5 & \underline{37.1} & 46.0 & \underline{29.0} & \textbf{47.7} & \underline{32.5} & 39.1 & \underline{23.8} & \textbf{43.7} & \underline{28.4} & 36.4 & \underline{21.0} \\ 
    \midrule
    FIERY~\cite{Hu2021ICCV} & \textbf{58.8} & 35.8 & \underline{50.5} & \underline{29.0} & \underline{47.4} & 30.1 & \underline{40.6} & 23.6 & \underline{41.8} & 26.7 & \underline{36.6} & 20.9 \\  
    StretchBEV-P &  \underline{58.1} & \textbf{52.5} & \textbf{53.0} & \textbf{47.5} & 46.8 & \textbf{32.7} & \textbf{43.7} & \textbf{38.4} & 38.2 & \textbf{31.8} & \textbf{37.4} & \textbf{30.8} \\
    \bottomrule
    \end{tabular}
    \caption{\textbf{Evaluation over Different Temporal Horizons.} This table extends Figure 3 in the main paper with the results of our models in comparison to FIERY \cite{Hu2021ICCV} over short~($2.0$s), mid~($4.0$s), and long~($6.0$s) temporal horizons.}
    \label{tab:long-temp}
\end{table}
In \tabref{tab:long-temp}, we provide the quantitative results of Fig. 3 in the main paper, which shows the performance comparison over different temporal horizons. As can be seen from the table, our models StretchBEV and StretchBEV-P outperform FIERY \cite{Hu2021ICCV} in far regions, especially StretchBEV-P by a large margin in terms of VPQ. The performance of StretchBEV is impressive in near IoU in longer settings.

In \tabref{tab:ged}, we provide Generalized Energy Distance (GED) for both of our models StretchBEV and StretchBEV-P compared to FIERY~\cite{Hu2021ICCV}. As can be seen from the table, both our models are more diverse than FIERY in terms of GED. This shows the significance of modelling future uncertainty with time-independent stochastic latent variables.

\boldparagraph{Run-time and Parameter Comparison} 
We compare the inference speed of our model StretchBEV-P and FIERY \cite{Hu2021ICCV} by measuring the average time needed to process a validation example in inference over 250 forward passes. Both models have almost the same inference speed (FIERY: $0.6436$ seconds/example vs. StretchBEV-P: $0.6469$ seconds/example). Although our model processes each time step separately, it does not introduce any drawbacks in terms of speed and its inference speed is almost the same as FIERY.

FIERY~\cite{Hu2018CVPR} has 8.1M parameters whereas StretchBEV-P has 16.2M. However, the runtime performances are still the same (0.64 sec per sample) thanks to the separation of generation from the learning of dynamics in our model.
\end{document}